\documentclass[journal]{IEEEtran}
\IEEEoverridecommandlockouts

\usepackage{microtype}
\usepackage{cite}

\ifCLASSINFOpdf
\usepackage[pdftex]{graphicx}
\DeclareGraphicsExtensions{.pdf,.jpeg,.png,.svg}
\else

\fi

\usepackage{amsmath}
\usepackage{amssymb}
\usepackage{amsfonts}
\usepackage{bm}
\usepackage{mathtools}
\usepackage{pifont}
\usepackage[dvipsnames]{xcolor}

\usepackage{enumitem}
\usepackage[ruled,vlined,linesnumbered]{algorithm2e}
\usepackage{algcompatible}
\SetAlCapNameFnt{\small}
\SetAlCapFnt{\small}
\SetAlCapSty{textsc}
\SetAlCapNameSty{textsc}

\usepackage{lettrine}
\usepackage{tikz}
\usetikzlibrary{positioning}

\usepackage[hidelinks]{hyperref}       %

\ifCLASSOPTIONcompsoc
\usepackage[caption=false,font=normalsize,labelfont=sf,textfont=sf]{subfig}
\else
\usepackage[caption=false,font=footnotesize]{subfig}
\fi

\usepackage[acronym]{glossaries}
\newacronym{GNN}{GNN}{Graph Neural Network}
\newacronym{CNN}{CNN}{Convolutional Neural Network}
\newacronym{RNN}{RNN}{Recurrent Neural Network}
\newacronym{MDN}{MDN}{Mixture Density Network}
\newacronym{GMM}{GMM}{Gaussian Mixture Model}
\newacronym{NLP}{NLP}{Natural Language Processing}
\newacronym{CVAE}{CVAE}{Conditional Variational Autoencoder}
\newacronym{GAN}{GAN}{Generative Adversarial Network}

\newacronym{GRU}{GRU}{Gated Recurrent Unit}
\newacronym{LSTM}{LSTM}{Long Short-Term Memory}
\newacronym{ELU}{ELU}{Expontential Linear Unit}

\newacronym{IVP}{IVP}{Initial Value Problem}
\newacronym{ODE}{ODE}{Ordinary Differential Equation}
\newacronym{NODE}{neural ODE}{Neural Ordinary Differential Equation}
\newacronym{EKF}{EKF}{Extended Kalman Filter}

\newacronym{EWTA}{EWTA}{Evolving Winner Takes All}
\newacronym{NLL}{NLL}{Negative Log-Likelihood}
\newacronym{SWA}{SWA}{Stochastic Weighted Averaging}
\newacronym{RMSE}{RMSE}{Root Mean Squared Error}
\newacronym{ADE}{ADE}{Average Displacement Error}
\newacronym{FDE}{FDE}{Final Displacement Error}
\newacronym{MR}{MR}{Miss Rate}
\newacronym{APDE}{APDE}{Average Path Displacement Error}
\newacronym{ANLL}{ANLL}{Average Negative Log-Likelihood}
\newacronym{FNLL}{FNLL}{Final Negative Log-Likelihood}

\newacronym{MPNN}{MPNN}{Message Passing Neural Network}
\newacronym{GAT}{GAT}{Graph Attention Network}
\newacronym{GCN}{GCN}{Graph Convolutional Network}

\newacronym{IA}{IA}{interaction-aware}

\newacronym{LCL}{LCL}{lane-change left}
\newacronym{LCR}{LCR}{lane-change right}
\newacronym{LK}{LK}{lane keep}

\newacronym{MDL}{MTP-GO}{}

\newacronym{CA}{CA}{Constant Acceleration}
\newacronym{CV}{CV}{Constant Velocity}
\newacronym{S-LSTM}{S-LSTM}{Social LSTM}
\newacronym{CS-LSTM}{CS-LSTM}{Convolutional Social Pooling}
\newacronym{T-TF}{T-TF}{Trajectory Transformer}
\newacronym{S2S}{Seq2Seq}{Sequence to Sequence}
\newacronym{MM-TF}{MM-TF}{Multimodal Transformer}
\newacronym{GNN-RNN}{GNN-RNN}{Graph Recurrent Network}

\newcommand{\highd}{\emph{highD}}
\newcommand{\ind}{\emph{inD}}
\newcommand{\round}{\emph{rounD}}

\DeclareMathOperator*{\argmin}{argmin}
\DeclareMathOperator{\arctantwo}{arctan2}
\newcommand{\R}{\mathbb{R}}
\newcommand{\set}[1]{\left\{#1\right\}}
\newcommand{\setsize}[1]{\left|#1\right|}
\newcommand{\tuple}[1]{\left(#1\right)}
\newcommand\norm[1]{\lVert#1\rVert}

\newcommand{\cmark}{\ding{51}}%
\newcommand{\xmark}{\ding{55}}%

\definecolor{MyGreen}{HTML}{10A23B}
\definecolor{MyRed}{HTML}{DD1111}

\newcommand{\gcmark}{\color{MyGreen}\cmark}%
\newcommand{\rxmark}{\color{MyRed}\xmark}%

\newcommand{\mdl}{MTP-GO}
\newcommand{\tplusplus}{Trajectron++}

\newcommand{\mmtf}{mmTransformer}

\newcommand{\graph}{\mathcal{G}}
\newcommand{\node}{\mathcal{V}}
\newcommand{\edge}{\mathcal{E}}
\newcommand{\agent}{\nu}
\newcommand{\neigh}[1]{N(#1)}
\newcommand{\ineigh}[1]{\tilde{N}(#1)}

\newcommand{\feature}[2][\agent{}]{\bm{f}_{#2}^{#1}}

\newcommand{\pos}[2][\nu]{\bm{x}^{#1}_{#2}}

\newcommand{\stcov}[1]{\bm{P}_{#1}}
\newcommand{\stjac}[1]{\bm{F}_{#1}}
\newcommand{\inpjac}[1]{\bm{G}_{#1}}
\newcommand{\pnoise}[1]{\bm{w}_{#1}}
\newcommand{\pnoisem}[1]{\bm{Q}_{#1}}

\newcommand{\dt}{T_s}

\newcommand{\stateestim}[1]{\hat{\bm{x}}_{#1}}
\newcommand{\mix}[1]{\bm{\pi}^{#1}}
\newcommand{\mean}[1]{\stateestim{#1}} %

\newcommand{\inp}[1]{\bm{u}_{#1}}

\newcommand{\state}[1]{\bm{x}_{#1}}

\newcommand{\std}{\sigma}
\newcommand{\corr}{\rho}

\newcommand{\history}{\mathcal{H}}
\newcommand{\predtime}{t}
\newcommand{\predhistory}{t_h}
\newcommand{\predhrz}{t_f}
\newcommand{\predagents}{\node_{\predtime}}

\newcommand{\hidden}{\bm{h}}
\newcommand{\hiddendim}{d_h}
\newcommand{\featuredim}{d_f}
\newcommand{\statedim}{d_s}

\newcommand{\initrep}{\hidden_{\text{init}}}
\newcommand{\encrep}[2][\agent]{\hidden_{#2}^{#1}}
\newcommand{\grurepx}[2][\agent]{\bm{\kappa}_{#2,i}^{#1}}
\newcommand{\grureph}[2][\agent]{\bm{\xi}_{#2,i}^{#1}}

\newcommand{\gnnf}[2][\agent{}]{\text{GNN}_{f}\left(\feature[#1]{#2}, \set{\feature[\tau]{#2}}_{\tau \neq{} #1} \right)}
\newcommand{\gnnh}[2][\agent{}]{\text{GNN}_{h}\left(\encrep[#1]{#2}, \set{\encrep[\tau]{#2}}_{\tau \neq #1} \right)}
\newcommand{\updatedrep}{{\hidden{}'}^{\agent{}}}

\newcommand{\ew}[2]{e_{#1,#2}}
\newcommand{\gcnd}[1]{d'_{#1}}
\newcommand{\edist}[2]{d_{#1,#2}}
\newcommand{\bwedge}{\sigma_{e}}
\newcommand{\gattw}[2][\agent{}]{\tilde{\alpha}_{\agent{},#2}}

\newcommand{\decrep}[2][\agent]{\hidden_{#2}^{#1}}
\newcommand{\posemb}[2][\nu]{\bm{\tilde{x}}^{#1}_{#2}}
\newcommand{\attw}[3][\agent{}]{\alpha_{#2,#3}^{#1}}

\newcommand{\attsum}[2][\agent{}]{\bm{s}_{#2}^{#1}}
\newcommand{\decinput}[2][\agent{}]{\bm{\hat{f}}_{#2}^{#1}}
\newcommand{\gnndec}[2][\agent{}]{\text{GNN}_{\hat{f}}\left(\decinput{#2}, \set{\decinput[\tau]{#2}}_{\tau \neq{} #1} \right)}

\newcommand{\transpose}{\text{T}} %

\newcommand{\eq}[2][my_equation]{\begin{equation}\label{eq:#1}#2\end{equation}}
\newcommand{\al}[2][my_equation]{\begin{align}\label{eq:#1}#2\end{align}} 

\newcommand{\totalepochs}{\mathcal{T}}
\newcommand{\wtaepochs}{\totalepochs{}_\text{EWTA}}
\newcommand{\warmepochs}{\totalepochs{}_\text{warm}}

\newcommand{\lrelu}{\gamma}

\DeclareMathOperator*{\argmax}{argmax}

\let\originalleft\left
\let\originalright\right
\renewcommand{\left}{\mathopen{}\mathclose\bgroup\originalleft}
\renewcommand{\right}{\aftergroup\egroup\originalright}

\usepackage{float}
\usepackage{xargs}

\usepackage{cleveref}
\Crefname{figure}{Fig.}{Figs}
\crefname{figure}{Fig.}{Figs}
\crefname{algorithm}{Algorithm}{Algorithms}
\crefname{table}{Table}{Tables}
\crefname{section}{Section}{Sections}
\crefname{equation}{}{}

\usepackage{multirow}
\usepackage{booktabs}
\usepackage{url}

\begin{document}

		\title{MTP-GO: Graph-Based Probabilistic Multi-Agent Trajectory Prediction with Neural ODEs}

		\author{Theodor Westny\IEEEauthorrefmark{1}, Joel Oskarsson\IEEEauthorrefmark{2}, Bj\"orn Olofsson\IEEEauthorrefmark{3}\IEEEauthorrefmark{1}, and Erik Frisk\IEEEauthorrefmark{1}
			\thanks{\hspace{-1em}This research was supported by the Strategic Research Area at Linköping-Lund in Information Technology (ELLIIT),
			the Swedish Research Council via the project \emph{Handling Uncertainty in Machine Learning Systems} (contract number: 2020-04122),
			and the Wallenberg AI, Autonomous Systems and Software Program (WASP) funded by the Knut and Alice Wallenberg Foundation.}
			\thanks{\scriptsize\IEEEauthorrefmark{1}Department of Electrical Engineering,
				Linköping University, Sweden.}
			\thanks{\scriptsize\IEEEauthorrefmark{2}Department of Computer and Information Science,
				Linköping University, Sweden.}
			\thanks{\scriptsize\IEEEauthorrefmark{3}Department of Automatic Control,
				Lund University, Sweden.}
		}

		\markboth{Journal of \LaTeX\ Class Files}%
		{}
		\maketitle

		\begin{abstract}
			Enabling resilient autonomous motion planning requires robust predictions of surrounding road users' future behavior.
			In response to this need and the associated challenges, we introduce our model titled MTP-GO.
			The model encodes the scene using temporal graph neural networks to produce the inputs to an underlying motion model.
			The motion model is implemented using neural ordinary differential equations where the state-transition functions are learned with the rest of the model.
			Multimodal probabilistic predictions are obtained by combining the concept of mixture density networks and Kalman filtering.
			The results illustrate the predictive capabilities of the proposed model across various data sets, outperforming several state-of-the-art methods on a number of metrics.

		\end{abstract}
		\begin{IEEEkeywords}
			Trajectory prediction, Neural ODEs, Graph Neural Networks.
		\end{IEEEkeywords}

		\IEEEpeerreviewmaketitle

		\section{Introduction}
\IEEEPARstart{A}{utonomous} vehicles are rapidly becoming a real possibility---reaching degrees of maturity that have allowed for testing and deployment on selected public roads \cite{othman2021public}. 
Future progress toward the realization of fully self-driving vehicles still requires human-level social compliance, heavily dependent on the ability to accurately forecast the behavior of surrounding road users.
In light of the interconnected nature of traffic participants, in which the actions of one agent can significantly influence the decisions of others, the development of behavior prediction methods is crucial for achieving resilient autonomous motion planning \cite{brudigam2021collision, batkovic2021robust, zhou2022interaction}.

As new high-quality data sets continue to emerge and many vehicles already possess significant computing power resulting from vision-based system requirements, the potential for adopting data-driven behavior prediction is increasing.
The application of \glspl{GNN} for the considered problem has emerged as a promising approach, partly because of their strong relational inductive bias that facilitates reasoning about relationships within the problem domain~\cite{battaglia2018relational}.
Furthermore, their capacity for multi-agent forecasting is a natural consequence of representing road users as nodes in the graph, enabling simultaneous predictions for multiple targets.

Despite their adaptability and performance in various tasks, deep networks often lack interpretability compared to traditional state-estimation techniques.
Recent methods have presented encouraging results by having a deep network compute the inputs to an underlying motion model \cite{cui2020deep, salzmann2020trajectron, li2021spatio}, convincingly improving interpretability through physically feasible predictions. 
These approaches are sensible, motivated by the comprehensive literature on motion modeling \cite{li2003survey,paden2016survey}.
However, different road users (e.g., pedestrians, bicycles, and cars) exhibit distinct dynamics, potentially necessitating a collection of customized models.
A possible solution is to employ a more general class of methods, such as \glspl{NODE} \cite{chen2018neuralode}, to learn inherent differential constraints.

\begin{figure}[!t]
	\centering
	\includegraphics[width=\columnwidth]{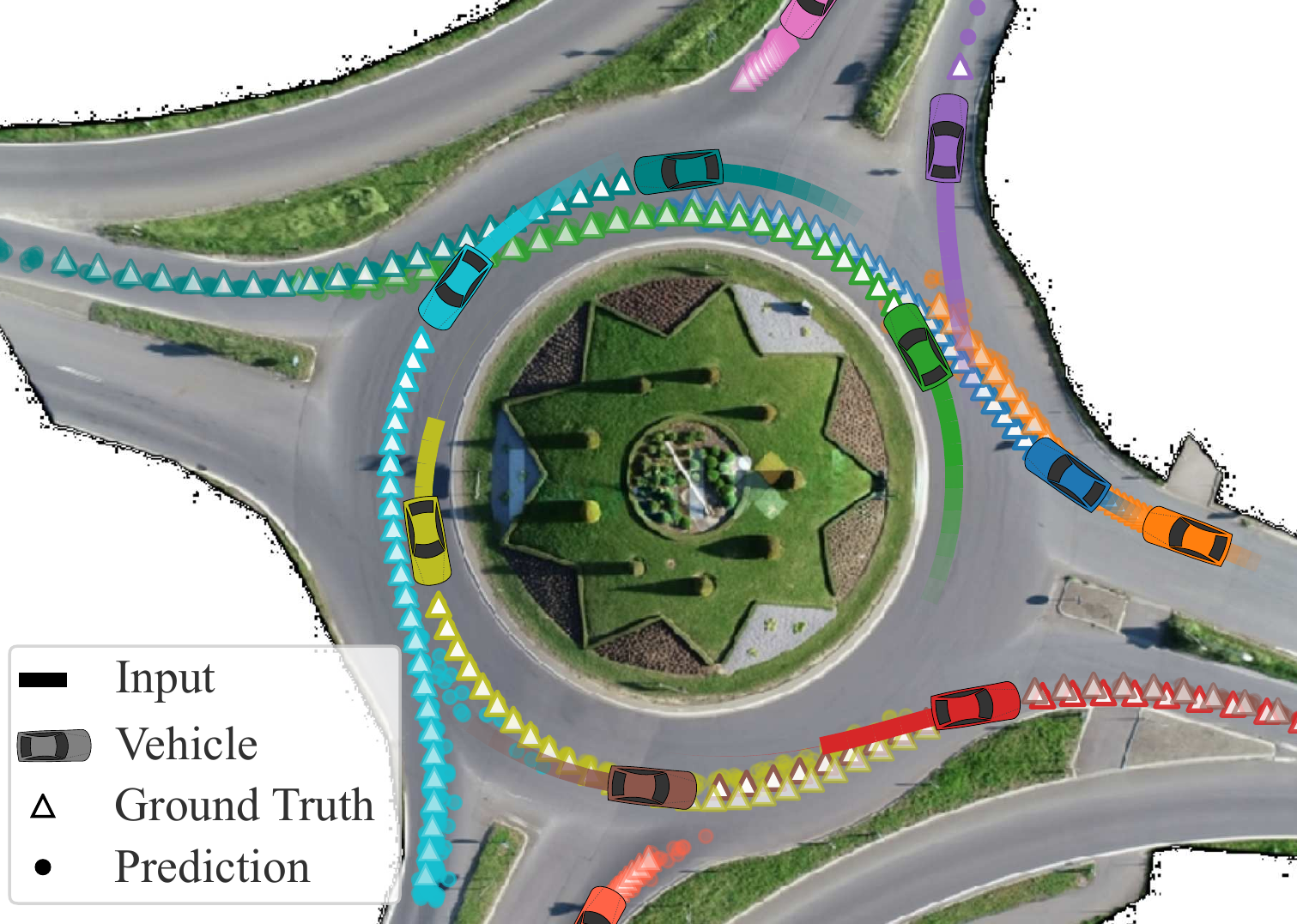}
	\caption{Example predictions by the proposed model, \mdl.
     Samples drawn from the learned distributions are here used to represent prediction uncertainty.
     The samples are closely aligned with the ground truth trajectories, illustrating the accuracy and confidence of the model.
     The example also displays the multimodal capabilities of the method.
     By studying the samples of the cyan-colored vehicle, two distinct predicted maneuvers can be seen.
     The image background is part of the \round{} data set~\cite{rounDdataset}.}
	\label{fig:round_pred}
    \vspace{-0.15in}
\end{figure}

Our main contribution is the proposed \mdl{}\footnote{\textbf{M}ulti-agent \textbf{T}rajectory \textbf{P}rediction by \textbf{G}raph-enhanced neural \textbf{O}DEs} model with key properties:
\begin{enumerate}[leftmargin=*]
    \item \textbf{Sustained relational awareness}: The model employs a spatio-temporal architecture using specially designed graph-gated recurrent cells that preserve salient inter-agent interactions throughout the complete prediction process.
    \item \textbf{Dynamic versatility}: The model is tailored to adapt to the perpetually dynamic environment while ensuring consistent forecasting for all agents present in the scene at prediction time, irrespective of possible historical information scarcity.
    \item \textbf{Naturalistic predictions}: To compute physically feasible trajectories, the model employs adaptable \glspl{NODE} to learn dynamic motion constraints from data, effectively capturing the inherently smooth nature of physical motion.
    \newline
    \item \textbf{Probabilistic forecasting}: To capture the inherent uncertainty and multifaceted nature of traffic, the model combines a mixture density output with an \gls{EKF} to compute multimodal probabilistic predictions.
\end{enumerate}
The \mdl{} model is compared to several state-of-the-art approaches and evaluated on naturalistic data from various traffic scenarios.   
Implementations are made publicly available\footnote{ \url{https://github.com/westny/mtp-go}}.

		\section{Related Work}
\label{sec:related_work}

\subsection{Traffic Behavior Prediction}
The topic of behavior prediction has attracted significant research interest over the past decade.
Comprehensive overviews can be found in surveys on the subject \cite{mozaffari2020deep, huang2022survey, fang2022behavioral}. 
Broadly, the area can be categorized into two primary streams, focusing on either \emph{intention} or \emph{motion} prediction.
Furthermore, a third category may also be recognized---predicting \emph{social patterns}, such as the driving style, attentiveness, or cooperativeness of human drivers \cite{alkinani2020detecting}. 
A majority of the methods use sequential input data, like historical agent positions, and often employ \glspl{RNN}, particularly \gls{LSTM} networks~\cite{mozaffari2020deep}, for their analyses.

The objective of intention prediction methods is to infer high-level decisions defined by the underlying traffic scene.
This includes predicting intention at intersections \cite{phillips2017generalizable}, or lane-change probability in highways \cite{lee2017convolution}.
Agent trajectories are labeled according to user-defined maneuvers, and models are trained using a supervised learning approach.
While intention prediction is a classification problem, the motion prediction task is regressive by nature, and distance-based measures are typically employed as learning objectives.
However, the two prediction problems are not disconnected.
This is illustrated in \cite{xin2018intention}, where the predicted intention is used to predict the future trajectory.
Because of the sequential nature of the motion prediction problem, several methods base their models on the encoder--decoder framework~\cite{sutskever2014sequence, cho2014learning}.

Numerous early studies in the field focused on single-agent motion prediction, taking into account only the historical observations of the individual target.
The concept of \emph{social pooling}, initially proposed for pedestrian trajectory prediction, was introduced in \cite{alahi2016social}---among the first works to demonstrate the effectiveness of \gls{IA} modeling.
The approach encodes interactions between neighboring agents using pooling tensors.
Building on the properties of \glspl{CNN}, the concept was extended in \cite{deo2018convolutional}, where the pooling layer was encoded using a \gls{CNN} to learn the spatial dependencies.
The model then uses an \gls{RNN}-based decoder to generate vehicle trajectories in highway settings.
In \cite{messaoud2020attention}, a more modular approach was proposed, wherein pairwise interactions between agents were learned using multi-head attention mechanisms.
Inspired by advancements in sequence prediction in other domains using \emph{Transformers} \cite{vaswani2017attention}, some researchers have explored their applicability to motion prediction. %
In \cite{giuliari2021transformer}, a Transformers-based architecture was proposed for the task of pedestrian trajectory prediction.
The idea was extended in \cite{liu2021multimodal, huang2022multi}, where IA-mechanisms and multimodal outputs were achieved using multiple Transformers.
Considering the challenges in formulating a supervised learning objective, other studies have investigated the use of inverse reinforcement learning to learn motion prediction tasks~\cite{fernando2020deep}.

Multimodality and probabilistic predictions are essential to capture the inherent uncertainty in traffic situations.
Several methods account for this, most commonly using \glspl{MDN} \cite{hu2018probabilistic, deo2018convolutional, messaoud2020attention}.
Other approaches include generative modeling, such as \gls{CVAE} \cite{salzmann2020trajectron, li2021spatio} and \gls{GAN} \cite{gupta2018social}.

A potential issue with black-box models is that they might output physically infeasible trajectories.
Given the considerable knowledge of motion modeling, it makes intuitive sense to use such knowledge also within data-driven models.
In response, some papers have recently included motion constraints within the prediction framework \cite{cui2020deep, salzmann2020trajectron, li2021spatio}.
Instead of employing pre-defined motion constraints that might be cumbersome to derive or be limited to a single type of road user, the \mdl{} model integrates neural ODEs to learn these constraints directly from the underlying data.

\subsection{Graph Neural Networks in Motion Prediction}
\glspl{GNN} is a family of deep learning models for graph-structured data \cite{wu2020comprehensive}.
Given a graph structure and a set of associated features, \glspl{GNN} can be used to learn representations of nodes, edges, or the entire graph \cite{gilmer2017neural}.
These representations can then be utilized in different prediction tasks.
\glspl{GNN} have demonstrated success in areas such as molecule generation \cite{zang2020moflow}, traffic flow prediction \cite{li2018diffusion}, and physics simulation \cite{rubanova2022constraint}.

When graph-structured data are collected over time, the resulting samples become time series with associated graphs.
Temporal \glspl{GNN} incorporate additional mechanisms to handle the time dimension.
These models can integrate \glspl{RNN} \cite{li2018diffusion}, \glspl{CNN} \cite{wu2019graph}, or attention mechanisms \cite{li2021spatio} to model temporal patterns. 
While most temporal \glspl{GNN} work with fixed and known graph structures \cite{li2018diffusion, yu2018spatiotemporal}, recent studies have also explored learning the graph itself \cite{Zhang2022graphguided}. 

\glspl{GNN} can be applied to trajectory prediction problems by letting edges in the graph represent interactions between entities or agents.
The LG-ODE model \cite{huang2020learning} uses an encoder--decoder architecture to predict trajectories of interacting physical objects.
A \gls{GNN} is used to encode historical observations, and a \gls{NODE} decoder predicts the future trajectories.
However, this model addresses general physical systems rather than focusing specifically on the traffic setting.

In \cite{diehl2019graph}, the use of different \glspl{GNN} for traffic participant interactions in motion prediction was investigated.
Although an early study on the topic, the research showed promising results for \gls{IA}-modeling.
GRIP++ \cite{li2019grip} is a graph-structured recurrent model tailored for vehicle trajectory prediction.
The scene is encoded to a latent representation using \gls{GNN} layers~\cite{kipf2017semisupervised}, which is then passed into an \gls{RNN}-based encoder--decoder network for trajectory prediction.
In \cite{jeon2020scale}, SCALE-net was proposed in order to handle any number of interacting agents.
Contrasting GRIP++, the node feature updates are encoded using an attention mechanism induced by graph edge features~\cite{gong2019exploiting}.
In \cite{mo2021graph}, node-wise interactions were proposed to be learned exclusively using a graph-attention mechanism~\cite{velickovic2017graph}.
The graph encoding is then passed to an LSTM-based decoder for vehicle trajectory prediction.
\tplusplus{}~\cite{salzmann2020trajectron} is a \gls{GNN}-based method that performs trajectory prediction using a generative model combining an \gls{RNN} with hard-coded kinematic constraints.
Similar to our method, it encodes a traffic situation as a sequence of graphs.
STG-DAT \cite{li2021spatio} is a similarly structured model to that of \tplusplus{}, and both are considered to be closely related to our model since they utilize temporal \glspl{GNN} to encode interactions and differentially-constrained motion models to compute the output.
Importantly, all above mentioned related works that combine GNNs with some recurrent module only consider the graph during the encoding stages.
Motivated by the importance of interaction-aware features, the proposed \mdl{} model instead maintains the graph throughout the entire prediction process, ensuring the retention of key interactions for the full extent of the prediction.

Some research has explored incorporating semantic information in the graph~\cite{zipfl2022relation, hu2022scenario}.
In \cite{hu2022scenario}, the graph is constructed based on semantic goals~\cite{hu2018probabilistic}, as opposed to being solely based on the agents themselves.
Although this method is not explicitly tailored for trajectory prediction, it provides a more general representation of the scene, which could potentially enhance generalization by transferability.
However, the method requires extensive knowledge of map-based information, such as lane lines and signaling signs, which might not always be available, nor does it incorporate agent motion constraints. 

		\section{Problem Definition}
\label{sec:problem_formulation}
The trajectory prediction problem is formulated as estimating the probability distribution of the future positions $\pos{\predtime+1}, \dots, \pos{\predtime+\predhrz}$ over the prediction horizon $\predhrz$ of all agents $\agent \in \node_{\predtime}$ currently in the scene.
The model infers the conditional distribution
\eq[traj_pred_dist]{
	p\left(\set{\tuple{\pos{\predtime+1}, \dots, \pos{\predtime+\predhrz}}}_{\agent \in \node_{\predtime}} \middle| \history \right)
}
given the history $\history$.
The history consists of observations $\feature{i} \in \R^{\featuredim}$, such as previous planar positions and velocities from time $\predtime - \predhistory$ to $\predtime$.
Given our focus on the method's future applicability within a predictive controller, the research emphasizes architectural aspects and usability in autonomous motion planning.
As a result, we primarily explore lightweight information typically available from vehicle onboard sensors.
Based on the outlined problem, this research aims to develop a prediction model that integrates several key properties:
\begin{enumerate}
	\item The model is required to be interaction-aware, learning interactions from historical observations but also preserving them throughout the entire prediction process.
	The model should therefore predict the future state of the current traffic \emph{scene}, rather than the future states of individual agents independently.
	\item As urban traffic is characterized by dynamic environments in which various road users enter and exit the vicinity of the ego vehicle, the model must accommodate this variability and consistently predict future trajectories for \textit{all} agents present at the time of prediction.
	This means that the model might have varying amounts of data from different agents when making the prediction.
	\item Vehicle trajectories typically reside on a smooth manifold; thus, the model should compute predictions that are smooth and dynamically feasible.
	This will be achieved by integrating a dynamic motion model, which is not fixed to a specific type but must be adaptable to different types of road users, e.g., pedestrians, bicycles, or cars.
	\item Since predictions of intention and motion will be inherently uncertain and multimodal, the model must represent the prediction uncertainty.
\end{enumerate}
		\section{Graph-based Traffic Modeling}
\begin{figure}[!t]
	\centering
	\includegraphics[width=\columnwidth]{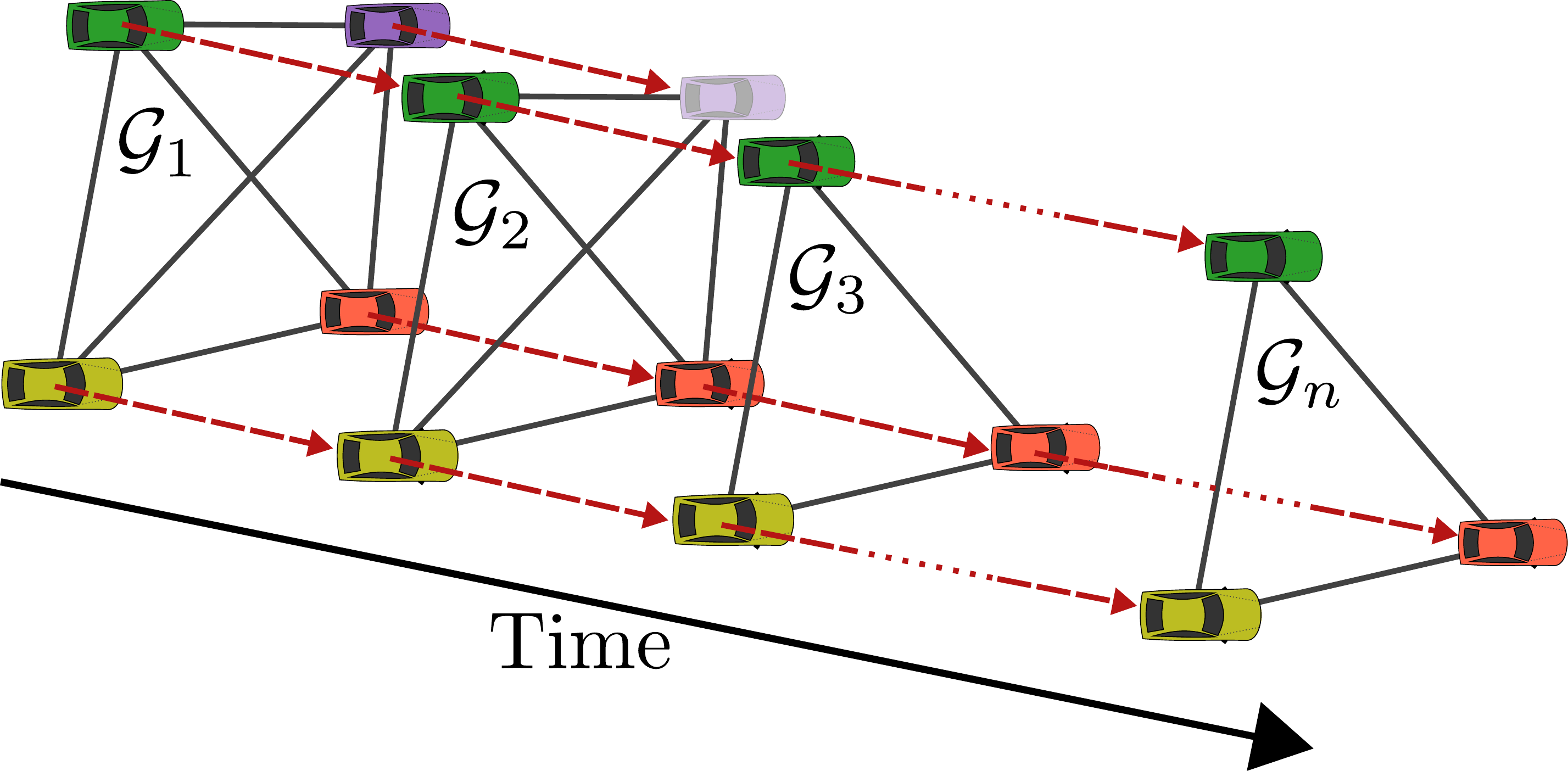}
    \caption{
        Example of a sequence of graphs describing a traffic situation starting with $\setsize{\node_1} = 4$ agents.
        After time 2, the faded agent leaves the traffic situation.
    }
	\label{fig:temporal_graph}
\end{figure}

\begin{figure}[!b]
	\centering
	\includegraphics[width=\columnwidth]{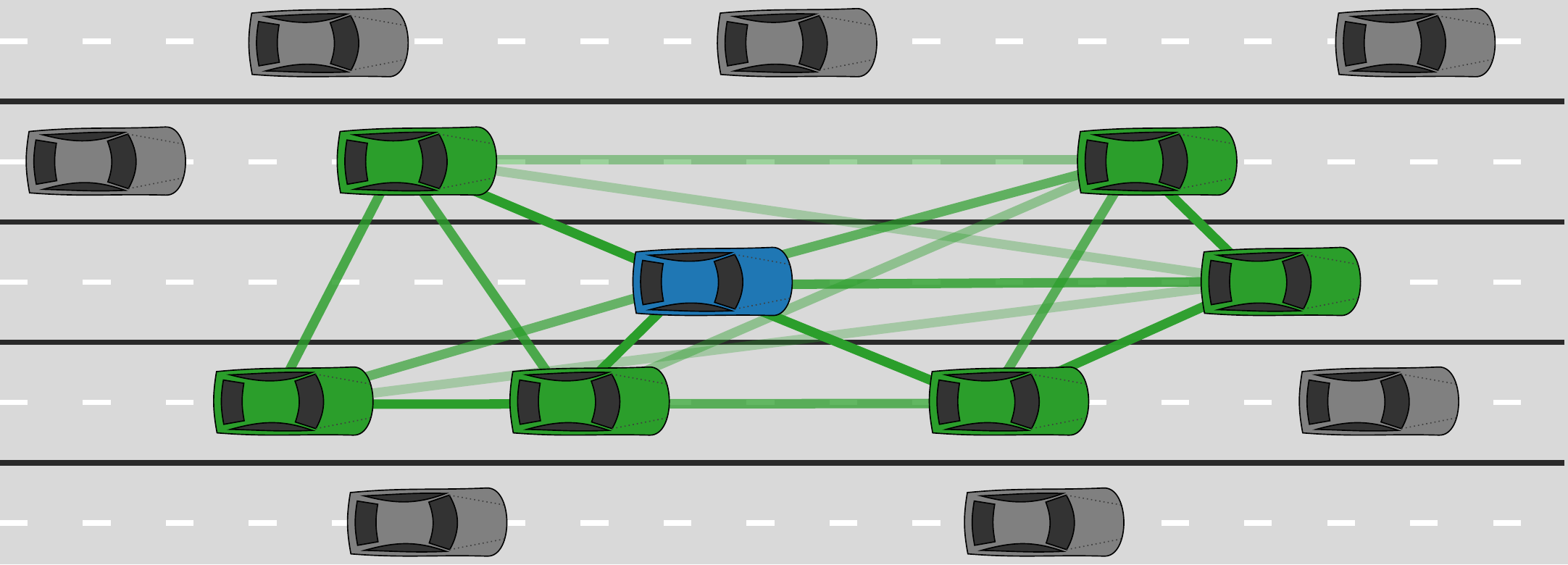}
	\caption{Graph construction for a highway forecasting problem, centered around a randomly selected vehicle.
	The graphs are complete with undirected edges and it is the model that learns the importance of each edge.
	This choice of structure greatly simplifies the implementation.}
	\label{fig:ego_graph}
\end{figure}

A traffic situation over $n$ time steps is modeled as a sequence $\graph_1, \dots, \graph_n$ of graphs.
The node set of the graph $\graph_i$ is $\node_i$, corresponding to the agents involved in the traffic situation.
The edge set $\edge_i$ is introduced to describe possible edge features.
As agents enter or leave the traffic situation over time, the graphs and the cardinality of the node sets change.
An example is shown in \cref{fig:temporal_graph}.
Trajectory forecasting is done for all agents $\node_{\predtime}$ in the traffic situation at time step $\predtime$.
The exact observation history of length $\predhistory$ can be summarized as
\eq[traffic_history]{
	\history = \left(
	\set{\graph_i}_{i=\predtime-\predhistory}^{\predtime},
	\set{\set{\feature{i}}_{\agent \in \node_i}}_{i=\predtime-\predhistory}^{\predtime}
	\right)
}
Graph construction follows an ego-centric approach, meaning that the input graph is built around a single vehicle; see \cref{fig:ego_graph}.
This is motivated by the connection to autonomous navigation, where predictions of surrounding vehicles are used for robust ego-vehicle decision-making.
Since agents can enter the traffic situation at time steps $i>\predtime-\predhistory$, the feature histories of different nodes can be of different lengths.
Regardless, the model should still output a prediction for all nodes in $\node_{\predtime}$, despite possible information scarcity.
Motivated by its connection to model predictive control \cite{fors2022resilient}, the prediction horizon was set to $\predhrz = 5$~s. 

\subsection{Input Features}
\label{sec:input_features}
The historic observations $\feature{i}$ associated with every node $\agent$ can be divided into \emph{node} and \emph{context} features.
These both refer to various positional information but are separated here since the context features are specific to each data set.
Both node and context features are time-varying by nature.
In addition, we make use of \emph{static} features, referring to node-specific time-invariant information.
Furthermore, edge features are also included in the form of Euclidean distance between the connected nodes.

\subsubsection{Node Features}
\label{sec:node_features}
The time-varying features pertaining to the nodes are summarized in \Cref{tab:node_feat}.
These include planar positions, velocities, and accelerations.
The planar coordinates are given relative to a pre-defined origin $(x_0, y_0)$, which is specific for each data set.
For highway scenarios, this refers to the graph-centered node's position at the prediction time instant.
For urban traffic scenarios, such as roundabouts and intersections, these refer to the roundabout circumcenter and the road junction point of intersection.

\begin{table}[!t]
	\caption{Node features}
	\label{tab:node_feat}
	\centering
	\begin{tabular}{c l l}
		\toprule
		Feature & Description & Unit\\
		\midrule
		$x$ & Longitudinal coordinate with respect to $x_0$ & m\\
		$y$ & Lateral coordinate with respect to $y_0$ & m\\
		$v_x$ & Instantaneous longitudinal velocity & m/s \\
		$v_y$ & Instantaneous lateral velocity & m/s\\
		$a_x$ & Instantaneous longitudinal acceleration & m/$\text{s}^2$ \\
		$a_y$ & Instantaneous lateral acceleration & m/$\text{s}^2$\\
		$\psi$ & Yaw angle  & rad \\
		\bottomrule
	\end{tabular}
\end{table}

\subsubsection{Road Context Features}
\label{sec:context_features}
For highway data, context features add additional information on lateral position with respect to the current lane centerline and the overall road center \cite{westny2021vehicle}.
The lane position, denoted $d_l$, refers to the vehicle's lateral deviation from the current lane center, bounded to the interval $[-1, 1]$ according to
\begin{equation}
	d_l = 2\frac{y+y_0 - l_{y,l}}{l_w}-1
\end{equation}
where $l_{y,l}$ is the lateral coordinate of the left lane divider for the current lane and $l_w$ is the lane width.
The road position $d_r$ is defined similarly, replacing the lane width with the breadth of the road and using the left-most lane divider as a reference.
The addition of these features is motivated by their potential to convey information about lane-changing maneuvers \cite{westny2021vehicle}.

For other scenarios, the context features are simply polar coordinates.
These are given with respect to the origin $(x_0, y_0)$ defined in \cref{sec:node_features} and computed according to:
\begin{subequations}
	\begin{align}
		r = \sqrt{(x_0 - x)^2 + (y_0 - y)^2}\\
		\theta = \arctantwo(y_0 - y, x-x_0).
	\end{align}
\end{subequations}

\subsubsection{Static Features}
\label{sec:static-features}
In this work, only the agent class (pedestrian, bicycle, car, bus, or truck) is considered for use as a static feature.
These are encoded using a \emph{one-hot} scheme and handled separately from the temporal features.
Static features are not included by default; a separate study on their effect and usability is presented in \cref{sec:static_experiment}.

		\section{Trajectory Prediction Model}
\label{sec:model}
\begin{figure*}[!t]
    \centering
    \includegraphics[width=\textwidth]{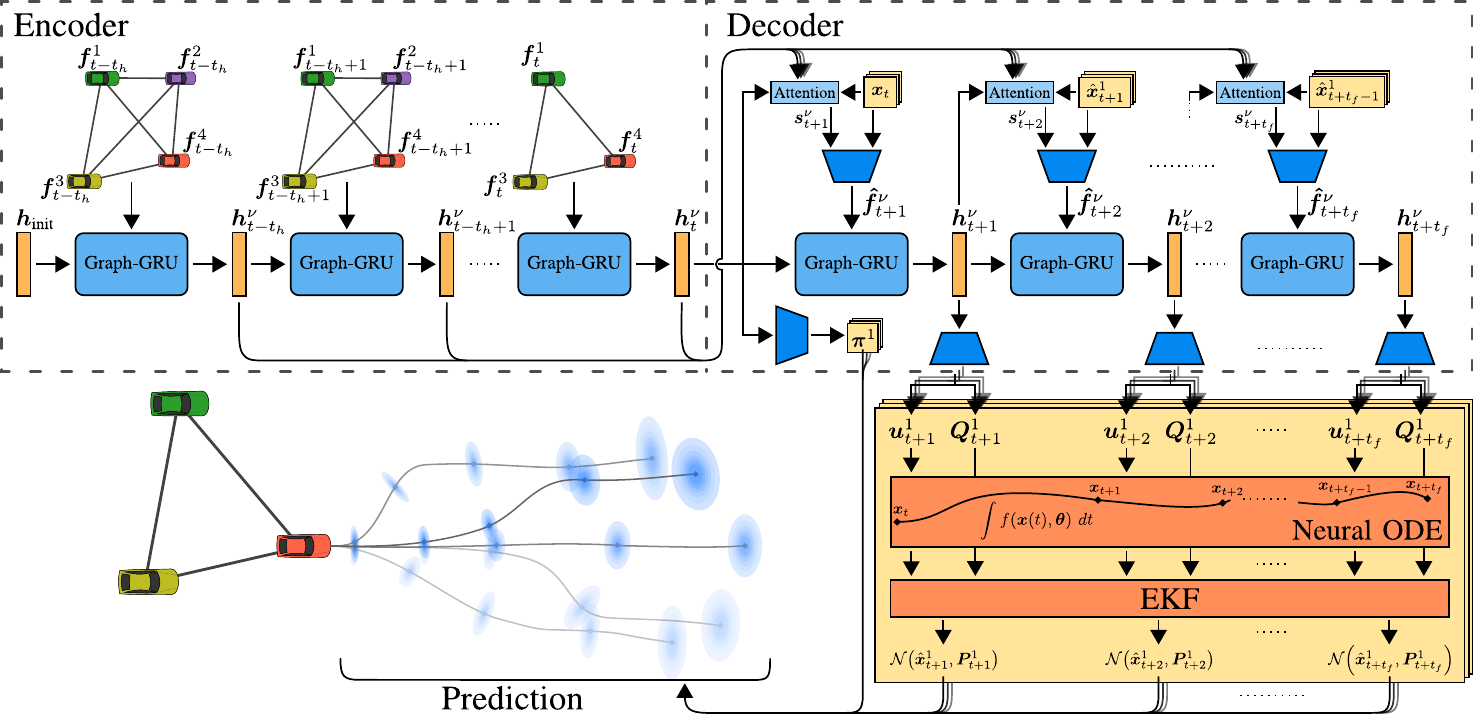}
    \caption{Schematic illustration of \mdl.
    The figure presents the process of computing predictions for a single agent $\agent{}$.
    The same process is performed concurrently for all agents in $\node_{\predtime}$.}
    \label{fig:model}
\end{figure*}

The complete \mdl{} model consists of a \gls{GNN}-based encoder--decoder module that computes the inputs to a motion model for trajectory forecasting.
The output is multimodal, consisting of several candidate trajectories $\stateestim{\predtime+1}^j, \dots, \stateestim{\predtime + \predhrz}^j$ for components $j\in\{1, \dots, M\}$.
In this paper, we set $M=8$, based on an assumption on the number of combined longitudinal and lateral modes in the data.
In addition, each candidate is accompanied by a predicted state covariance $\stcov{\predtime+1}^j, \dots, \stcov{\predtime + \predhrz}^j$ that is estimated using an \gls{EKF}.
\begin{itemize}
    \item \emph{Encoder}: The traffic scene history is encoded using a temporal \gls{GNN}. This module adopts an architecture based on \glspl{GRU} \cite{cho2014learning, cho2014properties} but replaces learnable weight matrices with graph neural networks.
	\item \emph{Decoder}: The decoder implements the same underlying structure as the encoder, with an added attention mechanism to learn temporal dependencies.
    The decoder computes the inputs to the motion model and estimates the process noise used in the \gls{EKF}.
    \item \emph{Motion model}: The motion model takes the output of the decoder and predicts the future states of the agents.
        The dynamics are modeled using \glspl{NODE} \cite{chen2018neuralode}.
\end{itemize}
A schematic of the full model is shown in \cref{fig:model}.

\subsection{Temporal \acrlong{GNN} Encoder}
The encoder takes the history $\history$ of all agents as input and computes a set of representation vectors useful for predicting future trajectories.
Using \glspl{GRU}, the history is processed sequentially to produce a set of $\hiddendim$-dimensional representations $\set{\encrep{i}}_{i=\predtime-\predhistory}^{\predtime}$ for each agent $\agent \in \predagents$.
A standard \gls{GRU} cell takes as input at time step $i$ the features $\feature{i}$ and the previous representation $\encrep{i-1}$.
Using these, six new intermediate vectors are computed according to
\begin{subequations}
\label{eq:gru_interm}
\al[gru_interm_x]{
    \left[\grurepx{r} \| \grurepx{z} \| \grurepx{h} \right] &= W_f \feature{i}\\
    \label{eq:gru_interm_f}
    \left[\grureph{r} \| \grureph{z} \| \grureph{h} \right] &= W_h \encrep{i-1}
}
\end{subequations}
where $W_f \in \R^{3\hiddendim \times \featuredim}$ and $W_h \in \R^{3\hiddendim \times \hiddendim}$ are learnable weight matrices, and $\|$ is the concatenation operation.
These vectors are then used to compute the representation $\encrep{i}$ for time step $i$ as
\begin{subequations}
\label{eq:full_gru_update}
\begin{align}
    \bm{r}_i^\agent &= \sigma(\grurepx{r} + \grureph{r} + \bm{b}_r) \\
    \bm{z}_i^\agent &= \sigma(\grurepx{z} + \grureph{z} + \bm{b}_z) \\
    \bm{\tilde{h}}_i^\agent &= \phi(\grurepx{h} + \bm{r}_i^\agent \odot \grureph{h} + \bm{b}_{h})\\
    \encrep{i} &= (\bm{1} - \bm{z}_i^\agent) \odot \bm{\tilde{h}}_i^\agent{} + \bm{z}_i^\agent \odot \encrep{i-1},
\end{align}
\end{subequations}
where the vectors $\bm{b}_r$, $\bm{b}_z$, $\bm{b}_h \in \R^\hiddendim$ are additional bias terms, $\odot$ is the Hadamard product, $\sigma$ is the sigmoid function, and $\phi$ is the hyperbolic tangent.
An initial representation $\initrep{}$ is used as input for the first encoding step.
This vector is learned jointly with other parameters in the model.
A standard \gls{GRU} \cite{cho2014properties} captures information in the history of agent $\agent$ but does not incorporate any observations from other agents. %

\subsubsection{Graph-Gated Recurrent Unit}
In order to accurately predict the future trajectories of each agent, it is important to consider inter-agent interactions.
Using the graph formulation in \mdl, such interactions can be captured by utilizing an extended \gls{GRU} cell where the linear mappings in \cref{eq:gru_interm} are replaced by \gls{GNN} components~\cite{zhao2018deep, oskarsson2022temporal}.
The \glspl{GNN} take as input not just the value at the specific node $\agent$ but also the values of other nodes in the graph.
The intermediate representations are computed by two \glspl{GNN} as
\begin{subequations}
\label{eq:gnn_interm}
\al[gnn_interm_f]{
    \left[\grurepx{r} \| \grurepx{z} \| \grurepx{h} \right]&=
        \gnnf{i}\\
    \label{eq:gnn_interm_h}
    \left[\grureph{r} \| \grureph{z} \| \grureph{h} \right] &=
        \gnnh{i-1}.
}
\end{subequations}
Note that since the \gls{GNN} components are built into the \gls{GRU} cell, the spatial properties of the problem are preserved throughout the encoding and decoding stages.
This contrasts \cite{li2019grip, jeon2020scale, mo2021graph, salzmann2020trajectron, li2021spatio} where the spatial encoding is done prior to the recurrent operations.

\subsubsection{\acrlong{GNN} Layers}
\label{sec:gnn_layers}
Each \gls{GNN} is built by multiple layers, each operating on all nodes concurrently.
Here the operation of each layer is described as centered on a node $\agent{}$.
Learnable parameters in the \glspl{GNN} are shared across all nodes but unique to each layer.
In this work, multiple types of \gls{GNN} layers from the literature are adopted:
\begin{itemize} %
    \item \emph{GraphConv}~\cite{morris2019weisfeiler} is a straightforward implementation of the \gls{MPNN} framework~\cite{gilmer2017neural}, which many \glspl{GNN} are based on.
        One GraphConv layer computes a new representation $\updatedrep{}$ of node $\agent{}$ according to
        \eq[graphconv_layer]{
            \updatedrep{} =
            \bm{b} +
            W_1 \hidden{}^\agent{} +
            \frac{1}{\setsize{\neigh{\agent{}}}}
                \smashoperator[r]{\sum_{\tau \in \neigh{\agent{}}}} \ew{\agent{}}{\tau} W_2 \hidden{}^\tau
        }
        where $W_1, W_2$, and $\bm{b}$ are learnable parameters, $\ew{\agent{}}{\tau}$ is a weight for the edge $(\agent{}, \tau)$, and $\neigh{\agent{}} = \{ \tau  \mid (\agent, \tau) \in \edge \}$ is the neighborhood of node $\agent{}$.
    \item \emph{\gls{GCN}}~\cite{kipf2017semisupervised} is motivated as a first-order approximation of a learnable spectral graph convolution.
        The \gls{GCN} layer update formulation for a node representation is
        \eq[gcn_layer]{
            \updatedrep{} =
            \bm{b} +
            \smashoperator[l]{\sum_{\tau \in \ineigh{\agent{}}}}
                \frac{\ew{\agent{}}{\tau}}{\sqrt{\gcnd{\agent{}} \gcnd{\tau}}} W_{2} \hidden{}^\tau
        }
        where $\ineigh{\agent{}} = \neigh{\agent{}} \cup \set{\agent{}}$ is the inclusive neighborhood of a node $\agent$ and $\gcnd{\agent{}} = 1 + \sum_{\tau \in \neigh{\agent{}}} \ew{\agent{}}{\tau}$.
    \item \emph{\gls{GAT}}~\cite{velickovic2017graph} layers use an attention mechanism to compute a set of aggregation weights over $\ineigh{\agent{}}$.
        This enables the \gls{GNN} to focus more on specific neighbors in the graph.
        The improved \gls{GAT}~\cite{brody2022attentative} version is used, in which attention weights are given by
        \eq[gat_attention]{
            \gattw{\tau} = \frac{
                \exp\left(\bm{a}_{\tilde{\alpha}}^\transpose{}
                \lrelu\left(
                    W_{\tilde{\alpha}} [\hidden{}^\agent{} \| \hidden^\tau \| \ew{\agent{}}{\tau}]
                \right)\right)
            }{
            \sum_{\upsilon \in \ineigh{\agent{}}}
                \exp\left(\bm{a}_{\tilde{\alpha}}^\transpose{}
                \lrelu\left(
                    W_{\tilde{\alpha}} [\hidden{}^\agent{} \| \hidden^{\upsilon} \| \ew{\agent{}}{\upsilon}]
                \right)\right)
            }
        }
        where $\bm{a}_{\tilde{\alpha}}$ and $W_{\tilde{\alpha}}$ are learnable parameters, and $\lrelu$ is the leaky ReLU activation function \cite{maas2013rectifier}. %
        Note that the edge weight $\ew{\agent{}}{\tau}$ is included as a feature in the computation of $\gattw{\tau}$.
        The new representation is then computed using the attention weights as
        \eq[gat_layer]{
            \updatedrep{} =
            \bm{b} +
            \smashoperator{\sum_{\tau \in \ineigh{\agent{}}}} \gattw{\tau} W_{2} \hidden{}^\tau.
        }
        \gls{GAT} layers can also use multiple \emph{attention heads}, which are independent copies of the attention mechanism described previously.
        Different heads can pay attention to different aspects, creating multiple separate representation vectors.
        These vectors are then concatenated or averaged in order to create a new representation.
    \item \emph{\gls{GAT}+} is a small extension of \gls{GAT} where an additional linear transformation is introduced only for the center node %
        \eq[gat_plus_layer]{
            \updatedrep{} =
            \bm{b} +
            W_1 \hidden{}^\agent{} +
            \smashoperator{\sum_{\tau \in \ineigh{\agent{}}}} \gattw{\tau} W_{2} \hidden{}^\tau.
        }
    This setup introduces additional flexibility in how the representation of the center node is used, something that has shown to be beneficial when evaluating on real data (see \cref{sec:results}).
\end{itemize}

\subsubsection{Gaussian Kernel Edge Weighting}
Following a commonly used method~\cite{yu2018spatiotemporal, li2018diffusion}, edge weights are computed using a Gaussian kernel:
\eq[edge_weight]{
    \ew{\agent{}}{\tau} = \exp\left(-\left(
        \frac{\edist{\agent{}}{\tau}}{\bwedge}
    \right)^2\right),
}
where $\edist{\agent{}}{\tau}$ is the Euclidean distance between agents $\agent{}$ and $\tau$.
The parameter $\bwedge$ controls how much to weigh down edges to agents that are far away.
This parameter is learned jointly with the rest of the model.

\subsection{Decoder with Temporal Attention Mechanism}
The decoder also uses a graph-based \gls{GRU} unit to compute the motion model inputs and process-noise estimates for time steps $\predtime+1, \dots, \predtime + \predhrz$.
As the true sequence of graphs is unknown for these time steps, the decoder always uses the last known graph $\graph_{\predtime}$ in the \gls{GRU}.
While the $\bm{h}$-input to these \gls{GRU} cells functions just as in \cref{eq:gnn_interm_h}, the $\bm{f}$-input is constructed through a temporal attention mechanism~\cite{vaswani2017attention}. %
At time $i > \predtime$, the attention weights $\{\attw{i}{l}\}_{l=\predtime-\predhistory}^{\predtime}$ are computed as
\begin{subequations}
    \begin{align}
    \bm{q}_i^\agent &= W_\alpha[\decrep{i-1} \| (W_x \pos{i-1} + \bm{b}_x)] + \bm{b}_\alpha \\
    \{\attw{i}{l}\}_{l=\predtime-\predhistory}^{\predtime} &= \left\{ \frac{\exp(q_{i, l}^\agent)}{\sum_{j=1}^{\predhistory} \exp(q_{i, j}^\agent)} \right\}_{l=1}^{\predhistory},
    \end{align}
\end{subequations}
where $W_x \in \R^{\hiddendim \times M\cdot\statedim}$, $\bm{b}_x \in \R^{\hiddendim}$, $W_\alpha \in \R^{\predhistory \times 2\hiddendim}$, and $\bm{b}_\alpha \in \R^{\predhistory}$ are learnable parameters.
For inference, the agent states $\pos{i-1}$ comes from the prediction of the $M$ components at the previous time step.
During training, teacher forcing \cite{goodfellow2016deep} is used, where the ground-truth value of $\pos{i-1}$ is used.
The attention weights represent how much attention is paid to the full encoder representation $\bm{o}^\agent_t = [\encrep{\predtime-\predhistory} \| \cdots \| \encrep{\predtime}]$ at decoder time step $i$.
The relevant encoded information is then summarized as
\eq[att_sum]{
    \attsum{i} = \sum_{l=\predtime-\predhistory}^{\predtime} \attw{i}{l} \encrep{l}
}
and the $\grurepx{\cdot}$-representations given by
\begin{subequations}
\al[dec_gru_input]{
    \decinput{i} &= \lrelu\left(W_{\hat{f}} [\attsum{i} \| \posemb{i-1}] + \bm{b}_{\hat{f}}\right)\\
    \left[\grurepx{r} \| \grurepx{z} \| \grurepx{h} \right] &=
        \gnndec{i}.
}
\end{subequations}
where $\lrelu$ is the leaky ReLU activation function.
The new decoder representation $\decrep{i}$ is finally computed according to \cref{eq:full_gru_update}.
This representation is then mapped through additional linear layers to compute the motion model input $\bm{u}$ and parameters defining the process noise matrix $\bm{Q}$ for all $M$ components.
Learnable parameters in the decoder are shared across all time steps.

\subsection{Motion Model}
\label{sec:motion_model}
There exist well-established models used in target tracking \cite{li2003survey} and predictive control applications \cite{paden2016survey} that hold potential for use in trajectory prediction.
However, while wheeled vehicles are often well described by nonholonomic constrained models, other road users, like pedestrians, may not share the same constraints.
The commonality between agents is that their motion is typically formulated mathematically using \acrshortpl{ODE}.
With the recent proposal of \glspl{NODE} \cite{chen2018neuralode}, such a flexible formulation could apply here by instead learning the underlying motion model but still enjoying the benefits of smooth trajectories.
A \gls{NODE} should learn the parameters $\bm{\theta}$ that best describe the state derivative
\begin{equation}
	\frac{d\bm{x}(t)}{dt} = f(\bm{x}(t), \bm{\theta}),
\end{equation}
where the states can be retrieved by solving an initial value problem.
This part of the network is denoted as the motion model $f$ and implemented using a fully-connected neural network with ELU activation functions \cite{clevert2015fast}.
The function $f$ accepts two input vectors: the prior state $\bm{x}$ and the current input $\bm{u}$.
Motivated by the degrees of freedom in ground vehicles, the dimensionality of the input $\bm{u}$ is fixed to two.
In \mdl, both a first and a second-order model are investigated.

\subsubsection{First-Order Model}
For the first-order model, there are two model states $(x, y)$.
Two separate \glspl{NODE}, $f_1$ and $f_2$, are used to describe the state dynamics where each model is associated with its respective input:
\begin{align}
    \begin{split}
    \dot{x} &= f_1(x, y, u_1) \\
    \dot{y} &= f_2(x, y, u_2)
    \end{split}
\end{align}

\subsubsection{Second-Order Model}
The same design principle is employed for higher-order models.
The \glspl{NODE} are assigned to model the highest-order state dynamics:
\begin{align}
    \begin{split}
    \dot{x} &= v_x \\
    \dot{y} &= v_y \\
    \dot{v}_x &= f_1(v_x, v_y, u_1) \\
    \dot{v}_y &= f_2(v_x, v_y, u_2)
    \end{split}
\end{align}
While it may not share the maneuverability properties of the first-order model, the additional integrations increase the smoothness of the trajectory, which might be useful in some scenarios.
For an extended study on motion models in graph-based trajectory prediction, see \cite{westny2023eval}.

\subsection{Uncertainty Propagation}
For multi-step forecasting, it is reasonable that the estimated uncertainty depends on prior predictions.
To provide such an estimate, the time-update step of the \gls{EKF} is employed.
For a given differentiable state-transition function $f$, input $\bm{u}_k$, and process noise $\bm{w}_k$:
\begin{equation}
	\label{eq:state_transition}
	\state{k+1} = f(\state{k}, \bm{u}_k) + \pnoise{k},
\end{equation}
the prediction step of the \gls{EKF} is formulated as:
\begin{subequations}
    \label{eq:ekf}
	\begin{align}
		\stateestim{k+1} &= f(\stateestim{k|k}, \inp{k}) \\
		\stcov{k+1} &= \stjac{k} \stcov{k|k} \stjac{k}^\transpose + \inpjac{k} \pnoisem{k|k} \inpjac{k}^\transpose
	\end{align}
\end{subequations}
where
\begin{equation}
	\stjac{k} = \frac{\partial f}{\partial \state{}}\Bigr|_{\stateestim{k|k}, \inp{k}}.
\end{equation}
In \cref{eq:ekf}, $\hat{\bm{x}}$ and $\bm{P}$ refer to the state estimate and state covariance estimate, respectively.
The matrix $\bm{Q}_k$ denotes the process noise covariance and should, in combination with $\bm{G}_k$, describe uncertainties in the state-transition function $f$, e.g., because of noisy inputs or modeling errors.
Here the state-transition function $f$ is taken as the current motion model.

To compute the estimated state covariance matrix $\stcov{k}$, the decoder is tasked with computing the covariance $\bm{Q}_k$ for every time step $k$.
For all motion models $f$, the process noise $\bm{w}$ is assumed to be a consequence of the predicted inputs and therefore enters into the two highest-order states.
Here, $\bm{w}$ is assumed to be zero-mean with covariance matrix
\begin{equation}
	\bm{Q} = \begin{pmatrix}
		\std_{1}^2 & \corr \std_{1} \std_{2} \\
		\corr \std_{1} \std_{2} & \std_{2}^2
	\end{pmatrix}
\end{equation}
where $-1\leq\corr\leq1$ and $\std_{1} > 0, \std_{2} > 0$. %
To get the correct signs of the correlation coefficient and standard deviations, they are passed through Softsign and Softplus activations \cite{goodfellow2016deep}.

How to construct $\inpjac{k}$ depends on the assumptions about the noise $\pnoise{k}$ and how it enters into $f$.
If it is assumed to be non-additive (cf. \cref{eq:state_transition}), such that $\state{k+1} = f(\state{k}, \inp{k}, \pnoise{k})$, then $\inpjac{k}$ is defined by the Jacobian:
\begin{equation}
	\inpjac{k} = \frac{\partial f}{\partial \pnoise{}}\Bigr|_{\stateestim{k|k}, \inp{k}}.
\end{equation}
While this is often a design choice, such a formulation is necessary for a completely general \gls{NODE} motion model, where, e.g., each input enters into every predicted state.
By the presented modeling approach and assuming the noise is additive, $\inpjac{k}$ can be designed as a matrix of constants.
In the simplest case with two state variables, then $\inpjac{k} = \dt \cdot I_{2}$, where $\dt$ is the sample time.
For higher-order state-space models, $\inpjac{k}$ can be generalized as:
\begin{equation}
	\inpjac{k}=  \dt
	\begin{pmatrix}
		 0 & 0 \\
		 \vdots & \vdots \\
		 1 & 0 \\
		 0 & 1
	\end{pmatrix}
\end{equation}

An interesting consequence is that this formulation simplifies generating $\stcov{}$ when the number of states $>2$.
In several previous works, only the bivariate case is considered. %
With the proposed approach, the method can be generalized to any number of states as long as $\inp{k}$ is 2-dimensional and $\pnoisem{k}$ is modeled explicitly.

\subsection{Multimodal Probabilistic Output}
\label{sec:probabilistic_output}
Inspired by \glspl{MDN}~\cite{bishop1994mixture, bishop2006pattern}, the model computes multimodal predictions by learning the parameters of a \gls{GMM} for each future time step.
Each output vector $\bm{y}_k$ of the model contains mixing coefficients $\mix{j}$, along with the predicted mean $\stateestim{k}^j$ of the states $\state{k}$ and estimated state covariance $\stcov{k}^j$ for all mixtures $j\in\{1, \dots, M\}$:
\begin{equation}
	\label{eq:gmm}
	\bm{y}_k = \left( \mix{j}, \big\{\stateestim{k}^j,  \stcov{k}^j\big\}_{j=1}^{M} \right).
\end{equation}
Note that the mixing coefficients $\mix{j}$, used to represent the weight of the component $j$, are constant over the prediction horizon $\predhrz$.
For notational convenience, the predictions will be indexed from $k=1, \dots, \predhrz$.
The learning objective is then formulated as the \gls{NLL} loss
\begin{equation}
	\label{eq:seq_loss}
	\mathcal{L}_{\text{NLL}} =\sum_{k= 1}^{\predhrz} -\log \left(\sum_j \mix{j}\mathcal{N}(\state{k} | \stateestim{k}^j, \stcov{k}^j)\right).
\end{equation}

\subsubsection{Winner Takes All}
\glspl{MDN} are notoriously difficult to train, and several attempts employ special training mechanisms such as learning parts of the distribution in sequence \cite{makansi2019overcoming}.
Although this might lead to increased stability in the learning process, many \glspl{MDN} still suffer from mode collapse.
The \gls{EWTA} loss, proposed in \cite{makansi2019overcoming}, is adopted for the prediction task to address these challenges:
\begin{subequations}
	\begin{align}
        \mathcal{L}_{\text{EWTA}}(K) &= \sum_{k=1}^{\predhrz}\sum_{j=1}^M c_{j} \ell(\mean{k}^j, \state{k}) \\
		c_{j} &= \delta (j \in B),
	\end{align}
\end{subequations}
where $\delta(\cdot)$ is the Kronecker delta and $B$ is the set of component indices pertaining to the current maximum number of winners $K$ (epoch dependent)
\begin{equation}
    B = \argmin_{\substack{A' \subset A\\ |A'|=K}} \sum_{k=1}^{\predhrz}\sum_{i \in A'}\ell(\mean{k}^i, \state{k}),
\end{equation}
where $A = \{1, \dots, M\}$.
This is applied as a loss function during the first training epochs.
For fast initial convergence, $\ell(\cdot)$ was chosen to be the \emph{Huber} loss function \cite{huber1992robust}---used for winner selection and training.
		\section{Evaluation \& Results}
\label{sec:results}
An evaluation of the capabilities of the proposed \mdl{} model was conducted using multiple investigations.
Based on the proposals in \cref{sec:gnn_layers}, a study on different \gls{GNN} layers and their usability for the considered task is discussed in \cref{seq:results-gnn}.
Second, an investigation of static features and their usability in the prediction context is presented in \cref{sec:static_experiment}.
Third, a comparison of the proposed model against related approaches across several data sets and scenarios is presented in \cref{sec:results-compare}.
Finally, the results of an ablation study are presented in \cref{sec:results-ablation}.

\subsection{Data Sets}
\label{sec:dataset}
Three different data sets: \highd{}~\cite{highDdataset}, \round{}~\cite{rounDdataset}, and \ind{}~\cite{inDdataset} were used for training and testing.
The data sets contain recorded trajectories from different locations in Germany, including various highways, roundabouts, and intersections.
The data contain several hours of naturalistic driving data recorded at 25 Hz.
The input and target data are down-sampled by a factor of 5, effectively setting the sampling time to $\dt=0.2$~s.
The maximum length of the observation window, i.e., the length of the model input, is set to $3$~s, motivated by prior work \cite{yoon2016layer}.
Because of the inherent maneuver imbalance, the \highd{} data were balanced prior to training using data re-sampling techniques.
This was done by oversampling lane-change instances and undersampling lane-keeping instances \cite{westny2021vehicle}.
The pre-processed \highd, \round, and \ind{} data sets consist of $100404$, $29248$, and $7820$ samples (graph sequences), respectively.
We allocate $80$\% of the total samples for training, $10$\% for validation, and $10$\% for testing.

\subsection{Training and Implementation Details}
All implementations were done in PyTorch \cite{paszke2019pytorch} and for \gls{GNN} components, PyTorch Geometric \cite{pyg2019} was used.
Jacobian computations were made efficient by the \texttt{\small functorch} package \cite{functorch2021}.
The optimizer \emph{Adam} \cite{kingma2014adam} was used with a batch size of 128.
Learning rate and hidden dimensionality were tuned using grid search independently for each experiment.

\subsubsection{Scheduling of Training Objective}
The model was trained using two loss functions, \gls{EWTA} and \gls{NLL}.
If $\totalepochs$ is the total number of training epochs, the \gls{EWTA} is used during an initial warm-up period of $\warmepochs = \totalepochs / 4$ epochs.
Additionally, for the first $\wtaepochs = \totalepochs / 8$ epochs the \gls{EWTA} loss is used exclusively.
The exact loss calculation for each epoch $n$ is described in \cref{alg:scheduling}.

\begin{algorithm}[!h]
	\caption{Scheduling of Training Objective}
	\label{alg:scheduling}
	\small
    \begin{algorithmic}[1]
        \IF{$n < \wtaepochs$}
            \STATE $K = \lceil M \cdot (\wtaepochs - n)/\wtaepochs \rceil$ //number of winners
            \STATE $\mathcal{L} = \mathcal{L}_{EWTA}(K)$
        \ELSIF{$\wtaepochs \leq n < \warmepochs$}
            \STATE $\beta = (\warmepochs - n)/(\warmepochs - \wtaepochs)$
            \STATE $\mathcal{L} = \beta \cdot\mathcal{L}_{EWTA}(K=1) + (1 - \beta)\cdot\mathcal{L}_{NLL}$
        \ELSE
            \STATE $\mathcal{L} = \mathcal{L}_{NLL}$
        \ENDIF
    \end{algorithmic}
\end{algorithm}

\subsection{Evaluation Metrics}
\label{sec:eval_metrics}
Typically, $L^2$-based metrics are used to evaluate prediction performance.
However, it is also important to measure the prediction likelihood, as it is an indicator of the model's ability to capture the uncertainty in its prediction.
The metrics are here presented for a single agent.
These values are then averaged over all agents in all traffic situations in the test set.
For the non-probabilistic metrics, it is important to consider which of the predicted components should be used.
Since \mdl{} is punished against mode-collapse during training, it is counter-intuitive to take the average over all components.
To that end, $L^2$-based metrics are provided with regard to $\stateestim{k}$ from the \gls{GMM} component $j^*$ with the predicted largest weight
\begin{equation}
	j^* = \argmax_j \mix{j}
\end{equation}
Here, the reduced state-vector $\state{}=[x, y]$ is used and correspondingly for $\stateestim{}$.

\begin{itemize}%
	\item \emph{\gls{ADE}}: The average $L^2$-norm over the complete prediction horizon is
	\begin{equation}
		\label{eq:ade}
		\text{ADE} = \frac{1}{\predhrz}\sum_{k=1}^{\predhrz} \norm{\stateestim{k} - \state{k}}_2
	\end{equation}
	\item \emph{\gls{FDE}}: 
	The $L^2$-norm of the final predicted position reflects the model's accuracy in forecasting distant future events:
	\begin{equation}
		\label{eq:fde}
		\text{FDE} = \norm{\stateestim{\predhrz} - \state{\predhrz}}_2
	\end{equation}
	\item \emph{\gls{MR}}: The ratio of cases where the predicted final position is not within $2$~m (from \cite{huang2022survey}) of the ground truth.
	This indicates prediction consistency. %
	\item \emph{\gls{APDE}}: The average minimum $L^2$-norm between the predicted positions and ground truth is used to estimate the path error.
	This is used to determine predicted maneuver accuracy:
	\begin{align}
		\label{eq:apde}
		\begin{split}
			\text{APDE} &= \frac{1}{\predhrz}\sum_{k=1}^{\predhrz} \norm{\stateestim{k} - \state{k^*}}_2 \\
			k^* &= \argmin_i \norm{\stateestim{k} - \state{i}}_2
		\end{split}
	\end{align}
	\item \emph{\gls{ANLL}}: This metric provides an estimate of how well the predicted distribution matches the observed data:
	\begin{equation}
		\text{ANLL} = \frac{1}{\predhrz}\sum_{k=1}^{\predhrz} -\log \left(\sum_j \mix{j}\mathcal{N}(\state{k} | \stateestim{k}^j, \stcov{k}^j)\right)
	\end{equation}
	It is also useful in determining the correctness of maneuver-based predictions if the method is multimodal.
	\item \emph{\gls{FNLL}}: The NLL equivalent of \gls{FDE} is
	\begin{equation}
		\text{FNLL} = -\log \left(\sum_j \mix{j}\mathcal{N}(\state{\predhrz} | \stateestim{\predhrz}^j, \stcov{\predhrz}^j)\right)
	\end{equation}
\end{itemize}

\subsection{Models Compared}
The following models are included in the comparative study:
\begin{itemize}[label={\scriptsize\raisebox{0.1ex}{\ding{69}}}]
	\item \gls{CA}: Open-loop model assuming constant acceleration.
	\item \gls{CV}: Open-loop model assuming constant velocity.
	\item \gls{S2S}: Baseline \gls{LSTM}-based encoder--decoder model (non-interaction-aware). %
	\item \gls{S-LSTM} \cite{alahi2016social}: Uses an encoder--decoder network  based on \gls{LSTM} for trajectory prediction.
	Interactions are encoded using social pooling tensors.
	\item \gls{CS-LSTM} \cite{deo2018convolutional}:
	Similar to \gls{S-LSTM}, but learns interactions using a \gls{CNN}.
	\item \gls{GNN-RNN} \cite{mo2021graph}: Encodes interactions using a graph network and generates trajectories with an \gls{RNN}-based encoder--decoder.
	\item \mmtf{} \cite{liu2021multimodal}: Transformer-based model for multimodal trajectory prediction.
	Interactions are encoded using multiple stacked Transformers.
	\item \tplusplus \cite{salzmann2020trajectron}: \gls{GNN}-based recurrent model.
	Performs trajectory prediction by a generative model together with hard-coded kinematic constraints.
	\item \mdl$1$: Our method with a first-order neural ODE.
	\item \mdl$2$: Our method with a second-order neural ODE.
\end{itemize}
The selection of related methods (\gls{S-LSTM}, \gls{CS-LSTM}, \gls{GNN-RNN}, \mmtf{}, and \tplusplus) was based on their relevance and the availability of the authors' code to the public.
Although the implementation of STG-DAT~\cite{li2021spatio}, a method closely related to ours, is not publicly accessible, we consider it comparable to \tplusplus{} due to their similarities.
In order to achieve a fair comparison, the methods were modified to make use of the same input features (see Sections~\ref{sec:node_features} through \ref{sec:context_features}) as \mdl{}, including edge weights for graph-based methods (\gls{GNN-RNN} and \tplusplus).
It was also observed that both \mmtf{} and \tplusplus{} were highly sensitive to feature scaling, requiring the standardization of input features to have zero mean and unit variance in order to attain desired performance.
Apart from these modifications, the models were preserved as per their original proposals and code.
All methods were subject to some hyperparameter tuning, specifically using methods to find good learning rates \cite{smith2017cyclical}, before being trained until convergence.

\subsection{Parameterizing \acrlong{GNN} Layers in \mdl}
\label{seq:results-gnn}
Different types of \gls{GNN} layers can be used in the graph components of \mdl.
Additionally, the \gls{GAT} and \gls{GAT}+ layers can incorporate different numbers of attention heads.
These choices can significantly impact the ability of the model to capture interactions in the traffic scenario.
In \cref{tab:gnn-res}, \mdl$2$ models with different types of \gls{GNN} layers are compared empirically on the \highd{} and \round{} data sets.
The metrics reported here are averaged over all vehicles in the traffic scene.

\begin{table}[!t]
    \caption{Performance of \mdl{}$2$ using different types of \gls{GNN} layers}
	\label{tab:gnn-res}
	\centering
    \resizebox{\columnwidth}{!}{%
	\begin{tabular}{l c c c c c c}
		\toprule
        & ADE & FDE & MR & APDE & ANLL & FNLL\\
		\midrule
        \highd{} &  & &  & & & \\
		\midrule
        GraphConv & $0.29$& $0.97$& $\bm{0.06}$& $0.28$& $-1.67$& $1.61$\\
        \gls{GCN} & $0.29$& $0.99$& $0.07$& $0.28$& $-1.75$& $1.60$\\
        \gls{GAT} (1 head) & $0.32$& $1.04$& $0.07$& $0.31$& $-1.56$& $1.68$\\
        \gls{GAT} (3 head) & $0.30$& $0.96$& $0.07$ & $0.28$& $-1.57$& $1.61$\\
        \gls{GAT} (5 head) & $0.28$& $0.91$& $\bm{0.06}$ & $0.27$& $-1.76$& $1.42$\\
        \gls{GAT}+ (1 head) & $0.29$& $0.94$& $\bm{0.06}$& $0.27$& $-1.75$& $1.43$\\
        \gls{GAT}+ (3 head) & $0.28$& $0.91$& $\bm{0.06}$& $0.27$&  $-1.78$& $1.44$\\
        \gls{GAT}+ (5 head) & $\bm{0.27}$& $\bm{0.90}$& $\bm{0.06}$& $\bm{0.26}$& $\bm{-1.86}$& $\bm{1.40}$\\
        \midrule
        \round{} &  & & & & & \\
		\midrule
        GraphConv & $1.03$& $3.28$& $0.38$& $0.62$& $-0.14$& $3.96$\\
        \gls{GCN} & $1.86$& $5.68$& $0.62$& $1.13$& $1.15$& $4.95$\\
        \gls{GAT} (1 head) & $1.36$& $4.12$&  $0.44$& $0.84$& $0.57$& $4.27$\\
        \gls{GAT} (3 heads) & $1.37$& $4.08$& $0.47$& $0.84$& $0.67$& $4.33$\\
        \gls{GAT} (5 heads) & $1.24$& $3.77$& $0.43$& $0.78$& $0.56$& $4.28$\\
        \gls{GAT}+ (1 head) & $\bm{0.97}$& $3.05$& $0.36$ &$\bm{0.60}$& $\bm{-0.17}$& $\bm{3.86}$\\
        \gls{GAT}+ (3 heads) & $0.98$& $3.06$& $\bm{0.35}$ & $0.61$& $-0.06$& $3.88$\\
        \gls{GAT}+ (5 heads) & $\bm{0.97}$& $\bm{3.02}$& $\bm{0.35}$& $0.62$& $-0.03$& $3.90$\\
		\bottomrule
	\end{tabular}}
\end{table}

The results indicate that the best choice for the \gls{GNN} layers in \mdl{} is \gls{GAT}+.
Also, the comparatively simple GraphConv layer performs surprisingly well.
Both of these layers feature some form of parameterization that handles the center node separately from the neighborhood.
This seems particularly useful for the roundabout scenario in \round{}.
In general, differences in performance are small for the \highd{} data set, but the choice of \gls{GNN} layer can be crucial for \round{}.
Using multiple attention heads is slightly beneficial for models with \gls{GAT} layers but does not make a notable difference for \gls{GAT}+.
All types of layers incorporate edge weights in some way.
Using these weights has shown to be important for accurately modeling agent interactions.
As a comparative example, a model with \gls{GCN} layers not using edge weights achieves an \gls{ADE} of $0.86$~m on \highd{} and $5.61$~m on \round{}.
On the \round{} data set in particular, the edge weights are highly informative, as there can be agents on the other side of the roundabout that do not impact the prediction significantly.
For comparison with other methods in the following sections, \mdl{} using \gls{GAT}+ layers with one attention head is used.

\subsection{Static Features in \mdl{}}
\label{sec:static_experiment}
\mdl{} makes no assumptions about the underlying motion model.
Therefore, it could potentially benefit from additional information about the types of agents present in the scene.
This was investigated by concatenating the neural-ODE inputs with a one-hot encoding of the agent classes (see \cref{sec:static-features}).
To fully illustrate their usability, the \ind{} data set was used because of its comprehensive content of diverse road users.
The resulting prediction performance is presented in \cref{tab:ind-res}, with the \gls{CA} and \gls{CV} models as references.
Here, the -S suffix adheres to models that include static features.
Note that the models are trained for all agent types concurrently; the results are only separated for the test data.

Overall, the first-order model, \mdl$1$, performs the best.
Interestingly, its connection to the \gls{CV} model, the best of the two reference models, illustrates that additional maneuverability is important in this context.
The addition of static features does not offer conclusive results.
While it does seem to have a minor positive impact on the second-order model, \mdl$2$, the results suggest the opposite for \mdl$1$. %
However, the most pronounced effect was instead observed during the training process, where models that included static features had a tendency to overfit toward the training data, becoming overconfident in their predictions.
Overall this indicates that the use of static features has a potential effect on prediction performance but requires additional research.
In the comparative study (\cref{sec:results-compare}), the static features are not included.

\begin{table}[!t]
	\caption{InD performance per road user}
	\label{tab:ind-res}
	\centering
	\begin{tabular}{l c c c c c c}
		\toprule
		Model & \multicolumn{2}{c}{\emph{Pedestrians}} & \multicolumn{2}{c}{\emph{Bicycles}} & \multicolumn{2}{c}{\emph{Cars}}  \vspace{0.05in}\\
		& ADE & FDE & ADE & FDE & ADE & FDE\\ 
		\midrule
		CA & $0.72$ & $2.20$ & $1.87$ & $5.95$ & $2.35$ & $7.61$ \\
		CV & $0.58$ & $1.42$ & $2.06$ & $5.40$ & $2.90$ & $7.42$\\
        \mdl$1$ & $\bm{0.38}$& $\bm{1.03}$& $\bm{0.90}$& $3.38$&  $\bm{1.07}$ & $3.38$\\
        \mdl$1$-S & $0.39$& $1.05$& $\bm{0.90}$& $\bm{2.65}$& $1.08$ & $\bm{3.32}$\\
        \mdl$2$ & $0.42$& $1.13$& $1.01$ & $3.00$ & $1.27$& $3.86$\\
        \mdl$2$-S & $0.40$& $1.08$& $1.00$ & $2.98$ & $1.20$& $3.69$\\
		\bottomrule
	\end{tabular}
\end{table}

\subsection{Comparative Study}
\label{sec:results-compare}
Since only a handful of the considered methods have multi-agent forecasting capabilities, the metrics are provided with regard to the graph-centered vehicle to provide a fair comparison.
The methods S-LSTM, CS-LSTM, and \mdl{} offer the possibility to compute the \gls{NLL} analytically, which is not the case for the sampling-based \tplusplus{}.
While it might not reflect the true likelihood, the NLL values of \tplusplus{} are computed using a kernel density estimate based on samples drawn from the predictive distribution \cite{salzmann2020trajectron} (marked by italics in \cref{tab:highd-res,tab:round-res}).
Furthermore, the non-probabilistic metrics of \tplusplus{} are computed by averaging over samples drawn from the most likely component.
Similarly, the metrics of \mmtf{} are computed based on the predicted trajectory with the largest confidence score~\cite{liu2021multimodal}.

\subsubsection{Highway}
\label{sec:results-compare-highway}
The performance on the \highd{} data set is presented in \cref{tab:highd-res}.
Given the high mean velocity of the vehicles in the data, the challenge of the task lies in predicting over large distances.
Despite this, the learning-based methods are very accurate, with a maximum reported FDE of approximately $1.7$~m; less than the length of an average car.
Out of all methods considered, \mdl{} shows the best performance across most metrics, except in reported NLL.
However, the estimated NLL of \tplusplus{} might not be comparable to the analytical NLL values of other models.
As a comparison, using the kernel density estimation method from \tplusplus{} to samples drawn from the predicted distribution of \mdl$2$, the calculated values are $\mathit{-7.16}$ and $\mathit{-3.35}$ for the average and final likelihood.
This illustrates a potential limitation of sampling-based estimates as they may not accurately represent the true likelihood of the distribution.
Although considered methodologically close to \mdl{}, \tplusplus{} does not achieve comparable performance on $L^2$-based metrics.
While the method was not originally intended for the investigated data set, it went through the same hyperparameter tuning as all others considered and trained approximately $4$ times longer, possibly attributed to difficulties linked to the \gls{CVAE} formulation \cite{salzmann2020trajectron}.
Still, \tplusplus{} achieves one of the lowest values on \gls{APDE}, indicating that it can predict the correct path, which is possibly attributed to the underlying motion model.
Between \mdl$1$ and \mdl$2$, the former achieves slightly better performance.
This is interesting considering that real highway driving is typically smooth, which would motivate the use of additional states in \mdl$2$.
Moreover, the \gls{CA} model performs relatively well compared to learning-based methods, possibly indicating that highway trajectory prediction is a simpler prediction problem.

\begin{table}[!t]
	\caption{HighD performance}
	\label{tab:highd-res}
	\centering
	\resizebox{\columnwidth}{!}{%
	\begin{tabular}{l c c c c c c}
		\toprule
		Model & ADE & FDE & MR & APDE & ANLL & FNLL\\
		\midrule
		CA & $0.78$ & $2.63$ & $0.55$ & $0.73$ & --- & ---\\
		CV & $1.49$ & $4.01$ & $0.79$ & $1.89$ & --- & ---\\
		Seq2Seq & $0.57$ & $1.68$ & $0.29$ & $0.54$ &  --- & ---\\ %
		S-LSTM~\cite{alahi2016social} & $0.41$ & $1.49$ & $0.22$ & $0.39$ & $-0.61$ & $3.20$\\ %
		CS-LSTM~\cite{deo2018convolutional} & $0.39$ & $1.38$ & $0.19$ & $0.37$ & $-0.66$ & $3.33$\\ %
		GNN-RNN~\cite{mo2021graph} & $0.40$ & $1.40$ & $0.17$ & $0.38$ & --- & ---\\
		\mmtf{}~\cite{liu2021multimodal} & $0.39$ & $1.13$ & $0.15$ & $0.39$ & --- & ---\\  
		\tplusplus{}~\cite{salzmann2020trajectron} & $0.44$ & $1.62$ & $0.23$ & $0.32$ & ($\mathit{-1.57}$) & ($\mathit{1.63})$\\ %
		\midrule
		\mdl$1$ & $\bm{0.30}$ & $\bm{1.07}$ & $\bm{0.13}$ & $\bm{0.30}$ & $\bm{-1.59}$ & $\bm{2.02}$\\ %
		\mdl$2$ & $0.35$ & $1.16$ & $0.15$ & $0.34$ & $-1.34$ & $2.18$\\ %
		\bottomrule
	\end{tabular}}
\end{table}

\begin{table}[!t]
	\caption{RounD performance}
	\label{tab:round-res}
	\centering
	\resizebox{\columnwidth}{!}{%
	\begin{tabular}{l c c c c c c}
		\toprule
		Model & ADE & FDE & MR & APDE & ANLL & FNLL\\
		\midrule
		CA & $4.83$ & $16.2$ & $0.95$ & $3.90$ & --- & ---\\ %
		CV & $6.49$ & $17.1$ & $0.94$ & $4.34$ & --- & ---\\ %
		Seq2Seq & $1.46$ & $3.66$ & $0.59$ & $0.82$ & --- & ---\\ %
		S-LSTM~\cite{alahi2016social} & $1.20$ & $3.47$ & $0.56$ & $0.74$ & $1.75$ & $5.12$\\  
		CS-LSTM~\cite{deo2018convolutional} & $1.19$ & $3.57$ & $0.60$ & $0.69$ & $2.09$ & $5.54$\\
		GNN-RNN~\cite{mo2021graph} & $1.13$ & $3.11$ & $0.51$ & $0.69$ & --- & ---\\
		\mmtf{}~\cite{liu2021multimodal} & $1.29$ & $3.50$ & $0.59$ & $0.77$ & --- & ---\\  
		\tplusplus{}~\cite{salzmann2020trajectron} & $1.09$ & $3.53$ & $0.54$ & $0.59$ & ($\mathit{-4.25}$) & ($\mathit{1.50}$)\\ 
		\midrule
		\mdl$1$ & $0.96$ & $\bm{2.95}$ & $\bm{0.46}$ & $0.59$ & $0.22$ & $\bm{3.38}$\\
		\mdl$2$ & $\bm{0.92}$ & $2.97$ & $0.48$ & $\bm{0.57}$ & $\bm{-0.22}$ & $3.85$\\ %
		\bottomrule
	\end{tabular}}
	\vspace{-0.1in}
\end{table}

\subsubsection{Roundabout}
\begin{figure}[!h]
	\centering
	\subfloat[While many methods achieve good prediction results, \mdl{} is the most accurate overall. In this example scenario, \mdl{} is the only method correct in its maneuver prediction.]{\includegraphics[width=\columnwidth]{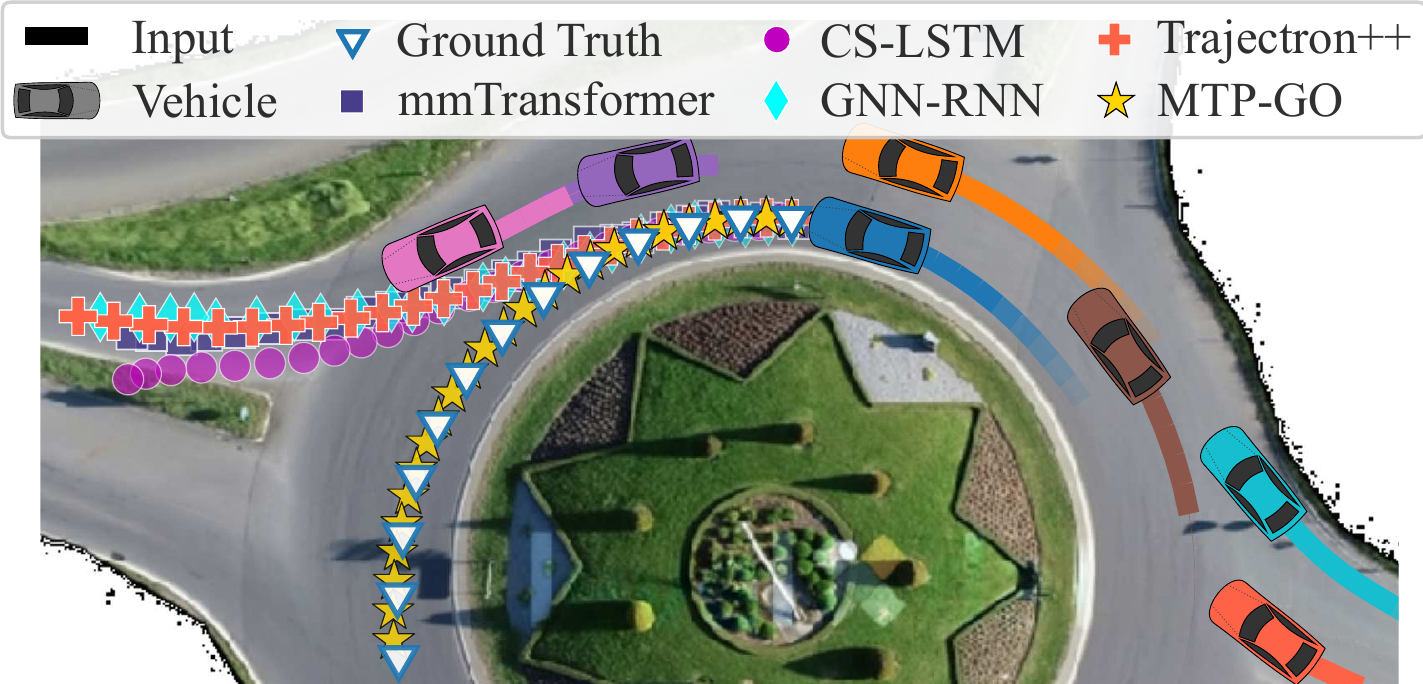}
	\label{fig:round-pred-circle}}
	\hfill
	\subfloat[Several methods struggle with predicting decelerating maneuvers.
	 Understanding the social queues without knowledge of traffic rules requires substantial interaction-aware capabilities and foresight.]{\includegraphics[width=\columnwidth]{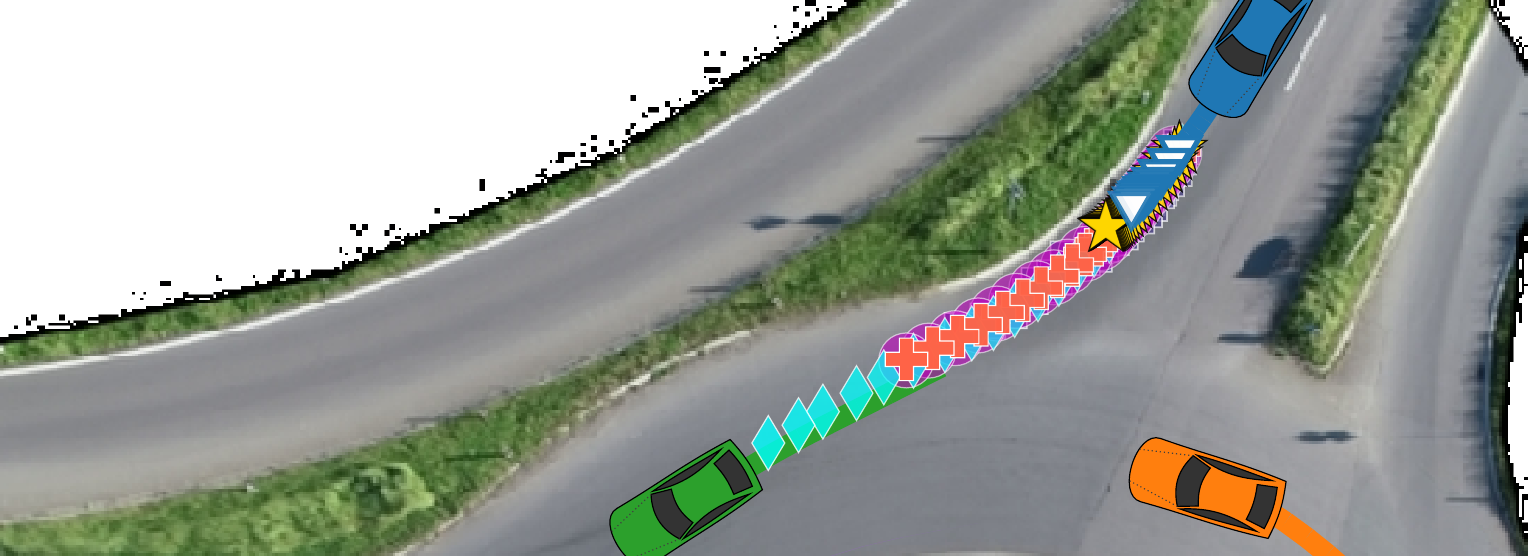}
	\label{fig:round-pred-acc}}
    \hfill
	\subfloat[Incorporating differential constraints into prediction models enhances their extrapolation capabilities.
	By extending the prediction horizon to $\predhrz=7$~s, the models must compute trajectories beyond the observed data (shown by the dashed line).
	Although this presents a challenge for the majority of models, \tplusplus{} and \mdl{} still perform well, which can be ascribed to the differential constraints within these models.]{\includegraphics[angle=90,width=\columnwidth]{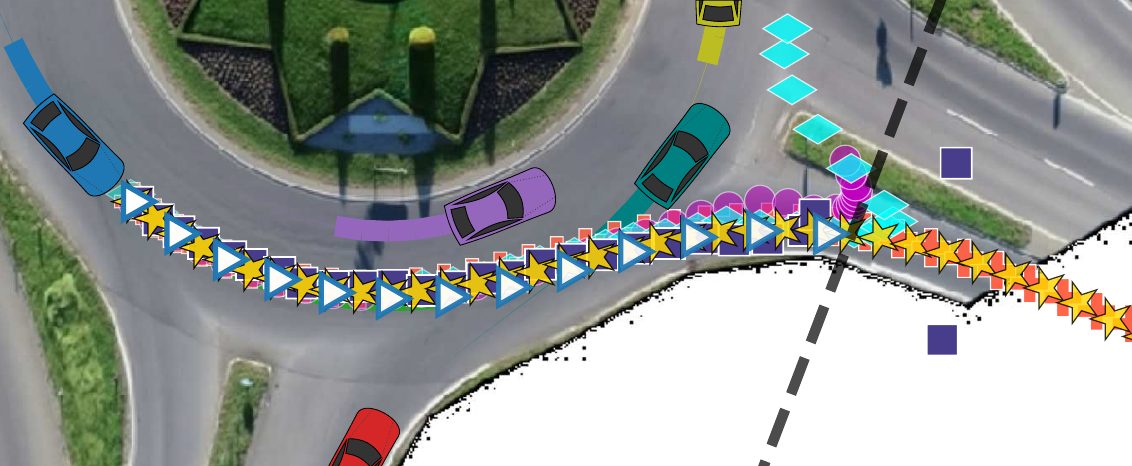}
	\label{fig:round-pred-best}}
	\caption{Example predictions from the roundabout test set using different compared models.
	 Scenarios have been specifically picked to illustrate the properties of the investigated methods.
	 For methods with multimodal capabilities, only the most probable component is shown for clarity.
	 The image background is part of the \round{} data set \cite{rounDdataset}.}
	\label{fig:round-pred}
	\vspace{-0.1in}
\end{figure}

In \cref{tab:round-res}, the performance is presented for the \round{} data set.
Upon initial examination, it is evident that predicting roundabout trajectories poses a greater challenge than the highway counterpart.
While the significance of interaction-aware modeling concerning the highway prediction problem was not so pronounced, the results are more indicative in this specific context, clearly favoring graph-modeled interactions.
This is further corroborated by the worse performance of the \gls{CA}, \gls{CV}, and \gls{S2S} models.
In \cref{fig:round-pred}, three different scenarios from the roundabout test set are illustrated.
Analyzing the prediction quality revealed that several related methods showed competitive performance.
However, one key contributor to the improved metrics of \mdl{} is its ability to correctly predict decelerating and yielding maneuvers, an aspect that many other methods struggle with (see \cref{fig:round-pred-acc}).
This is arguably attributed to the interaction-aware properties of \mdl{} and its capacity to preserve the graph.
Incorporating differential constraints within the prediction framework was observed to not only stabilize the training process but also provide valuable extrapolation capabilities, enabling the model to generalize beyond observed data. 
This presents a distinct advantage over conventional neural network models, which can exhibit shortcomings in this regard (see \cref{fig:round-pred-best}).

\vspace{-0.05in}
\subsection{Ablation Study}
\label{sec:results-ablation}
The \mdl{} model consists of several components, each with their contribution to the overall performance. 
In order to get a better understanding of how they affect the model's predictive abilities, an ablation study was conducted.
The goal of the ablation study is to dissect the architecture with the objective to determine which mechanisms are most important.
The study was conducted for both the \highd{} and \round{} data sets, using \mdl{}$2$ with single-attention-head \gls{GAT}+ layers.
Three components were considered for the ablation study:
\begin{enumerate}
	\item \emph{Graph Neural Network}: Removing the GNN is comparable to reducing the graph-GRU module to a standard GRU cell.
	In practice, this is done by removing all edges in the graphs, such that no nodes may share information. 
	This is investigated for both the encoder and decoder.
	\item \emph{Extended Kalman Filter}: Removing the EKF means that the decoder is tasked with directly computing the state covariance estimate.
	Although \mdl{}$2$ has four states, the covariance is only predicted for the first two, i.e., for a bivariate distribution over positions in $\mathbb{R}^2$.
	\item \emph{Neural ODE}: Eliminating the neural ODE removes differential constraints on outputs, requiring direct future state prediction. 
	Note that when the motion model is removed, the EKF cannot be used either (see (\ref{eq:ekf})), such that the complete model reduces to an MDN.
\end{enumerate}
To ensure that the outcomes were not resulting from random variations in model initialization and optimization, each model configuration was trained $10$ times using different random seeds.
Test performance is presented with an error margin, based on the $95\%$ confidence interval (CI)
\begin{equation}
	\text{CI} = \bar{X} \pm \text{t} \frac{S}{\sqrt{n}},
\end{equation}
where $\bar{X}$ is the sample mean, $S$ is the sample standard deviation, $n=10$ is the sample size, and $\text{t}=2.26$ derived from a t-distribution with $9$ degrees of freedom~\cite{walpole2012probability}. 
The metrics are computed based on an average of all agents in the scene, similar to the study in Section~\ref{seq:results-gnn}.
The results are presented in Table~\ref{tab:ablation} where a checkmark indicates a component's inclusion in the model whereas a cross indicates exclusion.

In comparison to nominal results, removing the differential constraints impairs the model's effectiveness, although it is more prominent for the highway scenario (H$_1$ vs. H$_5$).
Interestingly, when compared to the removal of other components which have more impact on $L^2$-based metrics, the likelihood is seemingly worst affected (R$_1$ vs. R$_5$). 
We hypothesize that this is because the model needs to predict a larger variance to accommodate for the added flexibility that comes with the removal of the differential constraints.

Removing the EKF also has a significant effect on the overall performance, seemingly on all metrics (see H$_4$ and R$_4$).
Although it should not directly affect the displacement error, the reason why performance on $L^2$-based metrics declines can be attributed to its impact on the learning process. 
By excluding the EKF, the model is tasked with directly computing the state covariance estimate while attempting to maximize the likelihood.
The increased difficulty of the task leads the model to concentrate more on refining covariance estimates rather than minimizing displacement errors.
This removal also makes the model more susceptible to random variations, evidenced by the variance in computed likelihood (see H$_4$ and H$_7$).

\begin{table*}[t!]
	\caption{Ablation Study}
	\label{tab:ablation}
	\centering
    \resizebox{\textwidth}{!}{%
	\begin{tabular}{l c c c c c | c c c c c c}
		\toprule
		Data Set & Index & Encoder GNN & Decoder GNN & EKF & ODE & ADE & FDE & MR & APDE & ANLL & FNLL\\
		\midrule
		 \multirow{7}{*}{\highd{}} & H$_1$ & \gcmark & \gcmark & \gcmark & \gcmark & $\bm{0.28 \pm 0.00}$ & $\bm{0.92 \pm 0.01}$ & $\bm{0.06 \pm 0.00}$ & $\bm{0.27 \pm 0.00}$ & $-1.74\pm0.05$ & $1.48\pm0.05$\\
         & H$_2$ & \rxmark & \gcmark & \gcmark & \gcmark & $\bm{0.28 \pm 0.00}$ & $\bm{0.92\pm 0.01}$ & $\bm{0.06 \pm 0.00}$ & $\bm{0.27 \pm 0.00}$ & $-1.75 \pm 0.06$ & $1.47 \pm 0.05$\\
         & H$_3$ & \gcmark & \rxmark & \gcmark & \gcmark & $\bm{0.28 \pm 0.00}$ & $0.94 \pm 0.01$ & $\bm{0.06 \pm 0.00}$ & $\bm{0.27 \pm 0.00}$ & $\bm{-1.81\pm 0.03}$ & $\bm{1.44 \pm 0.02}$\\
         & H$_4$ & \gcmark & \gcmark & \rxmark & \gcmark & $0.29 \pm 0.02$ & $0.97 \pm 0.03$ & $0.07 \pm 0.01$ & $0.28 \pm 0.02$ & $-1.76 \pm 0.86$ & $1.58 \pm 0.28$\\
         & H$_5$ & \gcmark & \gcmark & \rxmark & \rxmark & $0.44 \pm 0.02$ & $1.15 \pm 0.08$ & $0.18 \pm 0.03$ & $0.42 \pm 0.02$ & $0.63 \pm 0.12$ & $2.22 \pm 0.05$\\
         & H$_6$ & \rxmark & \rxmark & \gcmark & \gcmark & $0.30 \pm 0.00$ & $1.03 \pm 0.01$ & $0.07 \pm 0.00$ & $0.29 \pm 0.00$ & $-1.75 \pm 0.05$ & $1.61 \pm 0.02$\\
         & H$_7$ & \rxmark & \rxmark & \rxmark & \gcmark & $0.31 \pm 0.02$ & $1.07 \pm 0.03$ & $0.08 \pm 0.00$ & $0.30 \pm 0.01$ & $-1.34 \pm 1.12$ & $1.83 \pm 0.36$\\
  		 \midrule
           \multirow{7}{*}{\round{}} & R$_1$ & \gcmark & \gcmark & \gcmark & \gcmark & $0.99 \pm 0.02$ & $3.10 \pm 0.05$ & $\bm{0.37 \pm 0.02}$ & $\bm{0.61 \pm 0.02}$ & $-0.18 \pm 0.02$ & $\bm{3.78 \pm 0.02}$\\
           & R$_2$ & \rxmark & \gcmark & \gcmark & \gcmark & $\bm{0.98 \pm 0.01}$ & $\bm{3.05 \pm 0.04}$ & $\bm{0.37 \pm 0.02}$ & $\bm{0.61 \pm 0.02}$ & $\bm{-0.21 \pm 0.02}$ & $3.79 \pm 0.02$\\
           & R$_3$ & \gcmark & \rxmark & \gcmark & \gcmark & $1.07 \pm 0.02$ & $3.40 \pm 0.06$ & $0.40 \pm 0.02$ & $0.63 \pm 0.01$ & $-0.10 \pm 0.06$ & $3.92 \pm 0.03$\\
           & R$_4$ & \gcmark & \gcmark & \rxmark & \gcmark & $1.23 \pm 0.07$ & $3.83 \pm 0.18$ & $0.57 \pm 0.03$ & $0.72 \pm 0.03$ & $0.20 \pm 0.08$ & $4.08 \pm 0.07$\\
           & R$_5$ & \gcmark & \gcmark & \rxmark & \rxmark & $1.06 \pm 0.02$ & $3.05 \pm 0.05$ & $0.45 \pm 0.03$ & $0.65 \pm 0.02$ & $1.60 \pm 0.02$ & $3.48 \pm 0.01$\\
           & R$_6$ & \rxmark & \rxmark & \gcmark & \gcmark & $1.13 \pm 0.01$ & $3.64 \pm 0.04$ & $0.40 \pm 0.02$ & $\bm{0.61 \pm 0.02}$ & $-0.17 \pm 0.02$ & $3.95 \pm 0.02$\\
           & R$_7$ & \rxmark & \rxmark & \rxmark & \gcmark & $1.25 \pm 0.01$ & $3.98 \pm 0.04$ & $0.51 \pm 0.02$ & $0.70 \pm 0.01$ & $0.06 \pm 0.05$ & $4.03 \pm 0.03$\\
		\bottomrule
	\end{tabular}}
\end{table*}

This study, along with prior research, emphasizes the significance of modeling interactions for enhancing prediction accuracy. 
Notably, our work presents a novel contribution by exploring the placement of interaction-aware components within an encoder-decoder framework.
Interestingly, in the highway scenario, the placement of GNN components, whether situated in the encoder or decoder, does not appear to impact the results, provided they are present in some capacity (H$_6$ vs. $\{\text{H}_1, \text{H}_2, \text{H}_3\}$).
It is worth pointing out that eliminating the GNN entirely does not substantially diminish performance (H$_6$).
This observation is connected to the findings in Section~\ref{sec:results-compare-highway}, which reveal that although interaction-aware mechanisms enhance prediction performance, the improvement is less prominent in the highway study.
In contrast, the roundabout scenario reveals a different outcome---removing the GNN from the decoder leads to a considerable reduction in prediction performance across all metrics (R$_1$ vs. R$_3$).
However, eliminating the GNN from the encoder yields slight, albeit positive, improvements (R$_2$).
This is interesting for several reasons.
First, if excluding GNN operations from the encoder has minimal or no impact on performance, this translates to a reduced need for parameter optimization and decreased memory requirements. 
Secondly, this finding is in stark contrast to previous proposals in the context of behavior prediction, where GNN-based models typically incorporate graph operations exclusively during the encoding stages~\cite{li2019grip, jeon2020scale, mo2021graph, salzmann2020trajectron, li2021spatio}.
		\section{Conclusions}
In this paper, we have presented \mdl{}, a method for probabilistic multi-agent trajectory prediction using an encoder-decoder model based on temporal graph neural networks and neural ordinary differential equations.
By incorporating a mixture density network with the time-update step of an extended Kalman filter, the model computes multimodal probabilistic predictions.
Key contributions of \mdl{} include its interaction-aware capabilities, attributable to the model's ability to preserve the graph throughout the prediction process.
Additionally, the use of neural ODEs not only enables the model to learn the inherent differential constraints of various road users but also provides it with valuable extrapolation properties that enhance generalization beyond observed data.
\mdl{} was evaluated on several naturalistic traffic data sets, outperforming state-of-the-art methods across multiple performance metrics, and showcasing its potential in real-world traffic scenarios.

		\section*{Acknowledgment}
The authors would like to thank Fredrik Lindsten for helpful discussions and the anonymous reviewers for their valuable suggestions.
Computations were enabled by the supercomputing resource Berzelius provided by National Supercomputer Centre at Linköping University and the Knut and Alice Wallenberg foundation.

		\bibliographystyle{IEEEtran}
		\bibliography{IEEEabrv,references.bib}{}

	\end{document}